\documentclass[jornal]{IEEEtran}

\usepackage[utf8]{inputenc} 
\usepackage[T1]{fontenc}    
\usepackage{hyperref}       
\usepackage{url}            
\usepackage{booktabs}       
\usepackage{amsfonts}       
\usepackage{amsmath, bm}
\usepackage{mathrsfs} 
\usepackage{amsthm}
\usepackage{nicefrac}       
\usepackage{microtype}      
\usepackage{xcolor}         

\usepackage{graphicx}
\usepackage{caption}
\usepackage{subcaption}
\usepackage{enumerate}
\usepackage{cite}

\usepackage{tikz}
\usepackage{graphics}

\usepackage{amsfonts}
\usepackage{amsmath}
\usepackage{amssymb}
\usepackage{mathtools}
\usepackage{amsthm}
\usepackage{bbm}

\usepackage[capitalize,noabbrev]{cleveref}

\usepackage{algorithm}
\usepackage{algpseudocode}

\usepackage{enumitem}

\newtheorem{assumption}{\hspace{0pt}\bf Assumption}
\newtheorem{lemma}{\hspace{0pt}\bf Lemma}

\newtheorem{theorem}{\hspace{0pt}\bf Theorem}

\newcounter{excercise}
\newcounter{excercisepart}

\newcommand \red[1]         {{\color{red}#1}}

\newcommand \blue[1]        {{\color{blue}#1}}

\def \reals    {{\mathbb R}}

\def\mbE{{\ensuremath{\mathbb E}}}

\def\ccalD{{\ensuremath{\mathcal D}}}

\def\ccalF{{\ensuremath{\mathcal F}}}

\def\ccalI{{\ensuremath{\mathcal I}}}

\def\ccalL{{\ensuremath{\mathcal L}}}

\def\ccalN{{\ensuremath{\mathcal N}}}

\def\ccalU{{\ensuremath{\mathcal U}}}

\def\ccal0{{\ensuremath{\mathcal 0}}}
\def\bbA{{\ensuremath{\mathbf A}}}

\def\bbH{{\ensuremath{\mathbf H}}}
\def\bbI{{\ensuremath{\mathbf I}}}

\def\bbM{{\ensuremath{\mathbf M}}}

\def\bbP{{\ensuremath{\mathbf P}}}

\def\bbW{{\ensuremath{\mathbf W}}}

\def\bbS{{\ensuremath{\mathbf S}}}

\def\bbX{{\ensuremath{\mathbf X}}}

\def\bbb{{\ensuremath{\mathbf b}}}
\def\bbc{{\ensuremath{\mathbf c}}}

\def\bbf{{\ensuremath{\mathbf f}}}

\def\bbm{{\ensuremath{\mathbf m}}}

\def\bbp{{\ensuremath{\mathbf p}}}
\def\bbq{{\ensuremath{\mathbf q}}}
\def\bbr{{\ensuremath{\mathbf r}}}

\def\bbs{{\ensuremath{\mathbf s}}}

\def\bbx{{\ensuremath{\mathbf x}}}
\def\bby{{\ensuremath{\mathbf y}}}
\def\bbz{{\ensuremath{\mathbf z}}}
\def\bb0{{\ensuremath{\mathbf 0}}}
\def\rmB{{\ensuremath\text{B}}}

\def\rmD{{\ensuremath\text{D}}}

\def\rmP{{\ensuremath\text{P}}}

\def\rmh{{\ensuremath\text{h}}}

\def\tbx{{\widetilde{\ensuremath{\mathbf x}} }}

\def\bbepsilon{\boldsymbol{\epsilon}}

\def\bbtheta{\boldsymbol{\theta}}

\def\bblambda{\boldsymbol{\lambda}}

\def\bblam{\boldsymbol{\lambda}}
\def\bbmu{\boldsymbol{\mu}}
\def\bbnu{\boldsymbol{\nu}}
\def\bbxi{\boldsymbol{\xi}}

\def\bbTheta{\boldsymbol{\Theta}}

\def\bbPhi{\boldsymbol{\Phi}}

\def \bblam  {\bblambda}

\DeclarePairedDelimiterX{\infdivx}[2]{(}{)}{%
  #1\;\delimsize\|\;#2%
}

\newcommand{\pr}[1]{\mathbb{#1}}

\makeatletter
\DeclareRobustCommand{\cev}[1]{%
  {\mathpalette\do@cev{#1}}%
}
\newcommand{\do@cev}[2]{%
  \vbox{\offinterlineskip
    \sbox\z@{$\m@th#1 x$}%
    \ialign{##\cr
      \hidewidth\reflectbox{$\m@th#1\vec{}\mkern4mu$}\hidewidth\cr
      \noalign{\kern-\ht\z@}
      $\m@th#1#2$\cr
    }%
  }%
}

\makeatother

\title{Unrolled Neural Networks for
Constrained Optimization}
\author{Samar Hadou and Alejandro Ribeiro
\thanks{
Preliminary results of this work were presented at the ICASSP26 conference \cite{hadou2026unrolledgnns}.
The authors are with the Department of Electrical
and Systems Engineering, University of Pennsylvania, Philadelphia, PA 19104 USA (e-mail: selaraby@seas.upenn.edu; aribeiro@seas.upenn.edu).
}
}
\date{}

\begin{document}
%
\maketitle


\begin{abstract}
We develop unrolled neural networks to solve constrained optimization problems, offering accelerated, learnable counterparts to dual ascent (DA) algorithms. Our framework, termed constrained dual unrolling (CDU), comprises two coupled neural networks that jointly approximate the saddle point of the Lagrangian. The primal network emulates an iterative optimizer that finds a stationary point of the Lagrangian for a given dual multiplier, sampled from an unknown distribution. The dual network generates trajectories towards the optimal multipliers across its layers while querying the primal network at each layer. {Unlike standard unrolling, which maps the updates of a prescribed iterative solver directly to network layers, CDU preserves the underlying optimization dynamics while learning the update mappings. The desired primal–dual dynamics are therefore induced through constrained learning by enforcing primal-descent and dual-ascent constraints.} We formulate training the two networks as a nested optimization problem and propose an alternating procedure that updates the primal and dual networks in turn, mitigating uncertainty in the multiplier distribution required for primal network training. {We establish theoretical guarantees on the feasibility and optimality of the coupled networks.} We numerically evaluate the framework on quadratic programs (QPs) with linear constraints and power allocation in wireless networks. 
\end{abstract}


\section{Introduction}

Bridging the gap between model-based optimization and data-driven learning, algorithm unrolling embeds iterative solvers into neural network architectures, yielding learnable optimizers that are structured from domain-specific knowledge \cite{gregor_learning_2010, monga_algorithm_2021}. Beyond the interpretability inherited from standard optimizers, unrolling offers accelerated convergence, with unrolled models often achieving superior performance using substantially fewer layers (i.e., iterations) than conventional iterative algorithms. While the seminal work of Gregor and LeCun \cite{gregor_learning_2010}
introduced unrolling in the context of sparse coding, the approach has since achieved notable success in many applications such as computer vision \cite{zhang2020deep, Wei22, mou2022deep, Li20, qiao2023towards}, wireless networks \cite{hu2020iterative, chowdhury2021unfolding, Schynol23, Zhang2025}, and medical imaging \cite{wang2025proximal, li2021deep, chennakeshava2022deep, wang2023indudonet+}, among others \cite{hershey2014deep, nasser2022deep, Liu23, arab2025unrolled, Ravi2016OptimizationAA, hadou2026stochastic}.

Despite the extensive literature on unrolling, the vast majority of existing work focuses on unconstrained optimization problems. 
In the relatively few works that address constrained optimization, a common approach is to incorporate constraints into the training losses as penalty terms, effectively transforming the task into an unconstrained problem \cite{kishida2022temporal, bertocchi2020deep, hadou2026stochastic}. Another line of work considers unrolling primal-dual methods \cite{nagahama2022graph, Zhang2025, Yang2024, li2024pdhg, zhang2023joint}, but typically restricts learning to a small set of scalar hyperparameters, such as step sizes, rather than learning the full algorithmic dynamics. {For example,~\cite{zhang2023joint} unrolls a primal-dual procedure for joint precoding and CSI dimensionality reduction, while preserving the analytical primal-gradient and projected dual-ascent updates and learning parameters such as the associated step sizes.} As a result, these approaches tend to be narrowly tailored to the problem classes they were designed to address.

{In this paper, we introduce constrained dual unrolling (CDU), a framework for
constrained optimization that retains the iterative structure of primal-dual
methods while learning the update mappings themselves. Unlike traditional
algorithm unrolling, which explicitly maps the analytical updates of a
prescribed iterative solver to network layers, CDU uses neural networks to
represent the primal and dual iterates directly. The connection to the
underlying optimization algorithm is instead imposed through the training
constraints, which enforce descent of the primal residual and ascent of the
dual objective across the unrolled layers. Thus, CDU preserves the sequential
optimization structure of unrolling without requiring each layer to reproduce
a fixed analytical update rule.

CDU employs two coupled unrolled neural networks that collaboratively
approximate the saddle point of the Lagrangian. For a given multiplier, the
primal network predicts a stationary point of the Lagrangian, while the dual
network generates a sequence of multipliers toward the maximizer of the dual
function. We cast their training as a nested optimization problem, where the inner
level trains the primal network over a joint distribution of problem instances
and Lagrangian multipliers, while the outer level trains the dual network over
the distribution of problem instances.
A complication is that the multiplier distribution required by the primal
network is not known a priori, since it is induced by the trajectories
generated by the dual network. We therefore train the two networks
alternately. The dual network first generates multiplier trajectories used to
train the primal network, and the updated primal network is then used to train
the dual network. In this way, each network is adapted to the outputs of the other.

The use of learned update mappings distinguishes CDU from conventional
unrolling, but does not remove its iterative structure. Each unrolled layer
still maps one optimization iterate to the next and the sequence of layer
outputs forms a trajectory toward a solution. Similar departures from
explicitly prescribed updates have appeared in related areas, including
optimal control~\cite{kishida2022temporal}, deep equilibrium networks
\cite{bai2019deep}, and plug-and-play methods
\cite{pmlr-v97-ryu19a, graikos2022diffusion}. Such parameterizations are
motivated by the ability of neural networks to approximate continuous mappings,
including iterative update rules, under suitable regularity conditions
\cite{lu2017expressive, yarotsky2017error}.}

\begin{figure*}[t]
    \centering
    \includegraphics[width=\linewidth, height = 0.44\linewidth]{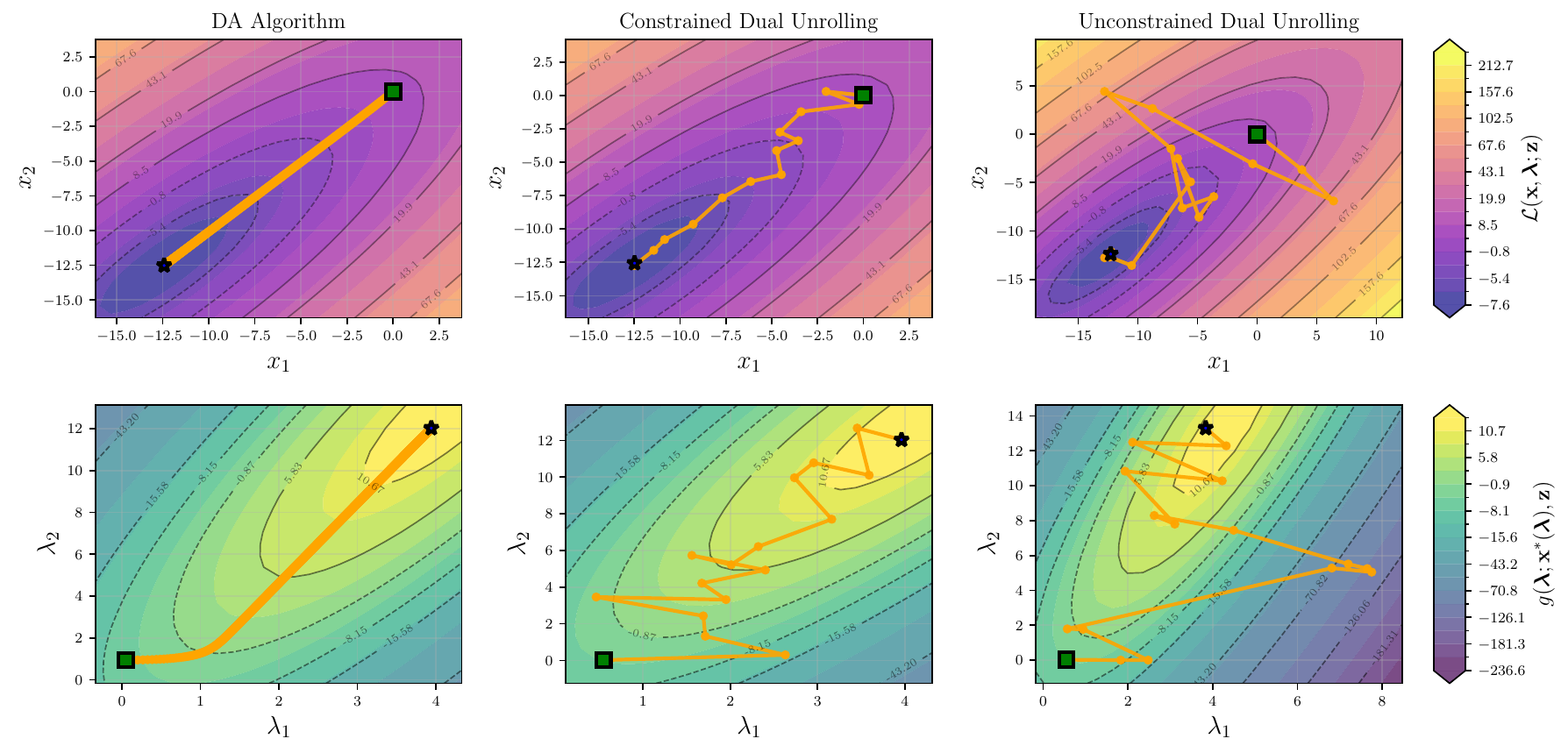}
      \caption{Trajectories generated by (left) DA algorithm, (middle) constrained dual unrolling and (right) its unconstrained counterpart for a QP instance: (Top) primal trajectories toward the stationary point of the Lagrangian $\ccalL(\cdot, \bblambda; \bbz)$, and (bottom) dual trajectories maximizing the dual function $g(\cdot \, ; \bbx^*(\bblambda),\bbz)$.
      The constrained networks generate descent trajectories emulating the DA algorithm, while the unconstrained models generate random trajectories and fail to hit the optimum of the dual function (bottom right).}
    \label{fig:comp2}
\end{figure*}

A second distinguishing aspect of CDU is the explicit enforcement of descent
and ascent behavior during training. Our prior work has shown that unrolled
models without such structure can be particularly susceptible to
out-of-distribution (OOD) examples~\cite{hadou2024Robust,porras2025constrained}. To address this
limitation, we impose monotonic descent and ascent constraints on the outputs
of the primal and dual layers, respectively.
Figure~\ref{fig:comp2} illustrates the effect of imposing these constraints on quadratic programs (QPs). The right panels show the evolution of the Lagrangian (top) and dual (bottom) functions across the layers of primal and dual networks.
The trajectories exhibit large excursions across the optimization spaces and fail to locate the maximum of the dual function as precisely as the iterative algorithm on the left. In contrast, the middle panels show that imposing the constraints yields descent and ascent trajectories toward the true solution with mild oscillations. Overall, our numerical experiments show that these descent constraints improve the final-layer performance and enhance the OOD generalization.

The most closely related work \cite{he2025primal} trains a graph neural network to imitate the trajectories of a greedy primal-dual algorithm designed for linear programs (LPs).
In contrast, our framework is agnostic to the primal and dual architectures and the specific constrained problem being solved. In this work, we tackle quadratic programs (QPs) with linear constraints and power allocation in wireless networks, using graph neural networks (GNNs). GNNs are particularly well-suited for these tasks, as they naturally capture the relational structure between variables and constraints, and are endowed with stability, transferability, and permutation equivariance \cite{Luana_graphon}. Nevertheless, our approach generalizes to a wide range of problems and architectures.

In summary, the main contributions of this work are:
\begin{itemize}
    \item We design a pair of primal and dual networks that interact at the layer level to solve the dual problem (Section~\ref{sec:architecture}).
    \item We impose descent and ascent constraints on the unrolled layers of the primal and dual networks and provide an alternating training scheme that updates the two networks in turn  (Section~\ref{sec:descent_constraints}).
    \item {We establish feasibility and optimality guarantees for the coupled networks, accounting for probabilistic layerwise descent and primal approximation errors (Section~\ref{sec:theory}).}
    \item We demonstrate the effectiveness of our approach on two different applications, QPs with linear constraints (Section \ref{sec:miqp}) and power allocation problems (Section~\ref{sec:pa}).
\end{itemize}

\section{Constrained-Optimization Unrolling}

Consider a constrained problem that poses the task of minimizing a scalar objective function $f_0: \mathbb{R}^n \to \mathbb{R}$, subject to $m$ constraints,
\begin{equation} \label{eq:problem}
    \begin{split}
        P^\star(\bbz) ~=~ \underset{\bbx \in \mathbb{R}^n}{\text{min}} \quad  f_0(\bbx;\bbz) \quad
        \text{s.t.}\quad  {\bf f}(\bbx; \bbz) \leq {\bf 0},
    \end{split}
\end{equation}
where ${\bf f}: \mathbb{R}^n \to \mathbb{R}^m$ is a vector-valued function representing the constraints, and $\bbz$ represents a problem instance. 
A solution $\bf x$ is \emph{feasible} if it satisfies the constraints and is considered \emph{optimal} if it minimizes the objective over all feasible solutions, i.e., ${\bf f}(\bbx; \bbz)\leq{\bf f}(\bby; \bbz)$ for all feasible $\bby$.
 
\subsection{Dual Formulations of Constrained Optimization}
We focus on dual methods, e.g., dual ascent (DA), which address \eqref{eq:problem} via its dual formulation. The Lagrangian function ${\cal L}: \mathbb{R}^n \times \mathbb{R}^m_+ \to \mathbb{R}$ combines the objective function with the constraints through Lagrangian multipliers $\boldsymbol{\lambda} \in \mathbb{R}_+^m$,
\begin{equation}
    {\ccalL}(\bbx, \boldsymbol{\lambda}; \bbz) ~=~ f_0(\bbx;{\bbz}) + \boldsymbol{\lambda}^\top {\bf f}(\bbx;{\bbz}).
\end{equation}
The dual problem is then defined as the max-min problem,
\begin{align}\label{eq:saddle-point}
    D^\star(\bbz) ~=~ \underset{\bblambda \in \pr{R}^m_+}{\text{max}} \, 
                \underset{\bbx}{\min} \quad \ccalL(\bbx, \bblambda; \bbz),
\end{align}
and the duality theory affirms that $D^\star(\bbz) \leq P^\star(\bbz)$ \cite{boyd2004convex}. The equality holds under strong duality, which applies to convex problems where both the objective $f_0$ and the feasible set defined by $\bbf$ are convex.
Under the strong-duality assumption, the Lagrangian has a saddle point $(\bbx^\star, \bblambda^\star)$, where $\bbx^\star$ is the minimizer of \eqref{eq:problem} and $\bblambda^\star$ is the maximizer of the dual problem. 
This motivates solving the dual problem, rather than \eqref{eq:problem}, in the case of convex problems and problems with provably small duality gaps, e.g., \cite{chamon2022constrained}.

Finding the saddle point of \eqref{eq:saddle-point} can be recast as the nested optimization problem,
\begin{align}
        D^\star(\bbz) ~=~ \underset{{\bblambda} \in \pr{R}^m_+}{\text{max}} \quad & g\big(\bblambda; \, \bbx^\star(\bblambda), \bbz\big) :=  \ccalL \big(\bbx^\star (\bblambda) , \bblambda; \bbz\big)\label{eq:outer} \\
    {\text{with} \quad \,} &
        {\bbx^\star}{(\bblambda)} \in \underset{\bbx}{\text{argmin}} \ \ccalL \big(\bbx, \bblambda; \bbz \big), \label{eq:inner}
\end{align}
where $g: \mathbb{R}^m_+ \to \mathbb{R}$ is the dual function.
The inner-level problem (Figure~\ref{fig:comp2}, top) finds the minimizer $\bbx^\star(\bblambda)$ of the Lagrangian function for any given $\bblambda$. The outer-level problem (Figure~\ref{fig:comp2}, bottom) maximizes the dual function $g$, which is defined as the Lagrangian function evaluated at the minimizer $\bbx^\star(\bblambda)$. 
We resort to the formulation in \eqref{eq:outer}--\eqref{eq:inner} because it mirrors the dynamics of the DA algorithm.

The DA algorithm, depicted in Figure~\ref{fig:comp2}~(left), computes the dual optimum $\bblambda^\star$ by repeating the following iteration,
\begin{align}
        \bbx_l & ~\in~ \underset{\bbx}{\text{argmin}} \ {\cal L}\big(\bbx, \bblambda_l; \bbz \big), \label{eq:DA_primal} \\
        \boldsymbol{\lambda}_{l+1} 
        & ~=~ \Big[ \, \boldsymbol{\lambda}_{l} + \eta \ \bbf \big( \bbx_l; \bbz \big) \, \Big]_+, \label{eq:DA_dual}
\end{align}
where $\eta$ is a stepsize, and $[\cdot]_+$ denotes a projection onto $\reals_+^m$. In each iteration, a Lagrangian stationary point $\bbx_l$ is attained for the current dual multiplier $\bblambda_l$, before a projected gradient ascent step is taken in the dual domain. The primal optimum $\bbx^\star$ is then recovered from the set $\bbx^\star(\bblambda^\star)$.

\subsection{Constrained-Optimization Unrolling}\label{sec:architecture}

We design a pair of unrolled neural networks that collaboratively find the saddle point of \eqref{eq:outer}--\eqref{eq:inner}. As illustrated in Figure~\ref{fig:architecture}~(middle), the dual network is the backbone of this design, where each dual layer evaluates the feasibility of the current iterate (cf. \eqref{eq:DA_dual}) and returns a new estimate of the dual variable that shapes the next primal iterate (cf. \eqref{eq:DA_primal}). Together, the primal and dual networks emulate the DA dynamics, as we describe next in more detail.

The primal network, denoted by $\bbPhi_\rmP(\, \bblambda, \bbz\, ;\bbtheta_\rmP)$ and illustrated in Figure~\ref{fig:architecture}~(left), predicts the minimizer of the Lagrangian function with respect to its first argument for a given multiplier $\bblambda$ and a problem instance $\bbz$. The network consists of a cascade of $K$ layers whose outputs constitute a trajectory from an initial point $\widetilde\bbx_0$ to a minimizer of \eqref{eq:inner}, i.e., $\widetilde\bbx_K \approx \bbx^\star(\bblambda)$. The $k$-th primal layer refines the estimate $\widetilde\bbx_{k-1}$ into
\begin{equation}\label{eq:primal_unrolling}
      \widetilde\bbx_k ~=~ \bbPhi_\rmP^k \Big(\widetilde\bbx_{k-1}, \bblambda, \bbz; \,\bbtheta_\rmP^k\Big),
\end{equation}
where $\bbtheta_\rmP^k$ is a learnable parameter, and $\bbtheta_\rmP = \{\bbtheta_\rmP^k \}_{k=1}^K$ collects all the parameters. We use the tilde notation to distinguish the intermediate estimates produced by the primal layers from other occurrences of $\bbx$ elsewhere in the architecture and analysis. The parametrization $\bbtheta_\rmP$ can vary depending on the optimization problem we solve, and the training approach we propose in this paper is agnostic to that choice. As a general design principle, an unrolled layer $\bbtheta^k_\rmP$ may consist of a multi-layer neural network, rather than a single layer, to leverage their expressivity. We also incorporate residual connections between unrolled layers as they naturally mimic gradient-based updates.

The dual network, denoted by $\bbPhi_\rmD(\, \bbz\,;\bbtheta_\rmD, \bbtheta_\rmP)$, approximates the maximizer of the dual function $g(\bblambda; \, \bbx^\star(\bblambda), \bbz)$. The dual function depends on $\bblambda$ directly and indirectly through $\bbx^\star(\bblambda)$, which is provided by the primal network. This induces the dependence of dual network $\bbPhi_\rmD$ on the primal parameterization $\bbtheta_\rmP$. The dual network has $L$ cascaded layers whose outputs trace a trajectory of estimates starting from an initial point $\bblambda_0$ and ending at an estimate of the optimal multiplier $\bblambda_L \approx \bblambda^\star$. 
The $l$-th dual layer is defined as
\begin{equation}\label{eq:dual_layer_unrolling}
    \bblambda_l ~=~ \bbPhi_\rmD^l\Big(\, \bblambda_{l-1}, \, \bbPhi_\rmP(\bblambda_{l-1}, \bbz; \bbtheta_\rmP), \, \bbz; \, \bbtheta_\rmD^l \, \Big),
\end{equation}
where $\bbtheta_\rmD^l$ is a learnable parameter and $\bbtheta_\rmD = \{\bbtheta_\rmD^l\}_{l=1}^L$. 
As per \eqref{eq:dual_layer_unrolling}, the $l$-th dual layer queries the primal network for its estimate $\bbx_{l-1} := \bbPhi_\rmP(\bblambda_{l-1}, \bbz; \bbtheta_\rmP)  \approx \bbx^\star(\bblambda_{l-1})$. Consequently, a single forward pass of the dual network triggers $L$ forward passes of the primal network, as illustrated in Figure~\ref{fig:architecture}~(middle). The design of the dual parameterization is problem-dependent, and we follow the same design principles as in the primal network. However, the nonlinearity at the end of each dual layer is forced to be a relu function to ensure that the predicted Lagrangian multipliers are always nonnegative.

Finally, the solution to \eqref{eq:problem}, $\bbx^\star \in \bbx^\star(\bblambda^\star)$, can be recovered by feeding the final dual estimate, $\bblambda_L = \bbPhi_\rmD\big(\bbz; \bbtheta_\rmD, \bbtheta_\rmP\big)$, to the primal network:
\begin{equation}\label{eq:recoverability}
    \bbx_L = \bbPhi_\rmP \Big(\, \bbPhi_\rmD\big(\bbz; \bbtheta_\rmD, \bbtheta_\rmP\big),\,  \bbz;\, 
    \bbtheta_\rmP \, \Big),
\end{equation}
as depicted in Figure~\ref{fig:architecture}~(right).
We distinguish the notation as follows: $\widetilde{\bbx}_k$ denotes the estimate generated by the $k$-th internal layer of the primal network, while $\bbx_l$ refers to the output of the primal network when queried with $\bblambda_l$ to approximate $\bbx^\star(\bblambda_l)$.



\definecolor{TRL9}{RGB}{93,128,62}
\pgfmathsetmacro{\colorsaturation}{50} 


\pgfmathsetmacro{\deltalayer}{-1.00} 
\pgfmathsetmacro{\deltashading}{0.350} 
\pgfmathsetmacro{\deltanetworks}{1.00} 
\pgfmathsetmacro{\deltalayeraux}{1.00} 

\pgfmathsetmacro{\nrlayers}{3} 

\pgfmathsetmacro{\xlabelshift}{ 0.1} 
\pgfmathsetmacro{\ylabelshift}{-0.1} 

\pgfmathsetmacro{\layerwidth}{3.2} 
\pgfmathsetmacro{\auxlayerwidth}{3} 


\tikzstyle{Arc} = [draw, -stealth]

\tikzstyle{arrow} = [thick, ->, >=stealth, draw]

\tikzstyle{block} = [ draw, 
                      fill = red!25, 
                      align = center,
                      inner sep = 0.3cm, 
                      anchor = north,
                      minimum height = 1.2cm, 
                      minimum width  = \layerwidth cm]

\tikzstyle{primalblock} = [ draw, 
                          fill = blue!15, 
                          align = center,
                          inner sep = 0.3cm, 
                          anchor = north,
                          minimum height = 1.2cm, 
                          minimum width  = \layerwidth cm]

\tikzstyle{optimalblock} = [ draw, 
                              fill = violet!15, 
                              align = center,
                              inner sep = 0.3cm, 
                              anchor = north,
                              minimum height = 1.2cm, 
                              minimum width  = \layerwidth cm]

\tikzstyle{block aux} = [ draw, 
                          fill = blue!8, 
                          align = center,
                          inner sep = 0.3cm, 
                          anchor = north,
                          minimum height = 1.2cm, 
                          minimum width  = \auxlayerwidth cm]

\tikzstyle{shade} = [draw, black!30, fill = red!5]
\tikzstyle{primalshade} = [draw, black!30, fill = blue!5]
\tikzstyle{optimalshade} = [draw, black!30, fill = violet!5]


{

\begin{figure*}[t]

\centering

\fontsize{7}{7}\selectfont

\begin{tikzpicture}[scale = 1.0]


    \pgfdeclarelayer{bg}     
    \pgfsetlayers{bg,main}   

    
    \path (0,0) node (layer0) {};

    \path (layer0) ++ (-2.0,0) node (lambdanode) {};
    \path (lambdanode) ++ (\xlabelshift+0.1,\ylabelshift-0.1) node [below right] {\red{$\bblambda_l$}};
        
    \foreach \i in {1,...,\nrlayers}
    {
        \pgfmathtruncatemacro{\imone}{\i-1} 
        
        \path (layer\imone.south) ++ (0,\deltalayer) node [primalblock] (layer\i) 
            {$\bbPhi_\rmP^{\i}\big(\, \tbx_\imone, \red{\bblambda_l}, \bbz; \bbtheta_\rmP^{\i} \, \big)$};    
        
        \path[arrow] (layer\imone.south)--(layer\i.north);
        \path        (layer\imone.south) ++ (\xlabelshift,\ylabelshift) node [below right] {$\tbx_\imone$};

        \path[arrow] (lambdanode.east) |- (layer\i.west);
    }   
    
    \path (layer\nrlayers.south) ++ (0,\deltalayer) node (output) {};
    
        \path[arrow] (layer\nrlayers.south)--(output);
        \path (output) ++ (\xlabelshift,-\ylabelshift) node [above right] 
            {$\blue{\tbx_\nrlayers} = \bbPhi_\rmP(\red{\bblambda_l}, \bbz; \bbtheta_\rmP)$};

    \begin{pgfonlayer}{bg} 

        \path [] (layer1.north west)          ++ (-\deltashading-0.2,  \deltashading) coordinate (nw);
        \path [] (layer\nrlayers.south east)  ++ ( \deltashading, -\deltashading) coordinate (se);
        
        \path [primalshade] (nw) -- (nw -| se) -- (se) -- (se -|nw) -- (nw);
                
    \end{pgfonlayer}{bg}

    
    \path (layer0) ++ (\layerwidth, 0) ++ (\deltanetworks, 0) node (layer0) {};
        
    \foreach \i in {1,...,\nrlayers}
    {
        \pgfmathtruncatemacro{\imone}{\i-1} 
        
        \path (layer\imone.south) ++ (0,\deltalayer) node [block] (layer\i) 
            {$\bbPhi_\rmD^{\i}\big(\, \red{\bblam_\imone}, \blue{\bbx_\imone}, \bbz; \bbtheta_\rmD^{\i} \, \big)$};    

        \path (layer\i.north east) ++ (\auxlayerwidth/2, 0) ++ (\deltalayeraux, 0) node [block aux] (layeraux\i)
            {$\bbPhi_\rmP(\red{\bblambda_\imone}, \bbz; \bbtheta_\rmP)$};    
        
        \path[arrow] (layer\imone.south)--(layer\i.north);
        \path        (layer\imone.south) ++ (\xlabelshift,\ylabelshift) node [below right] (lam\imone) {$\red{\bblam_\imone}$};

        \path[arrow] (layer\i.east) ++ (0, 0.1) -- ++ (\deltalayeraux, 0) ;
        \path[arrow] (layeraux\i.west) ++ (0, -0.1) -- ++ (-\deltalayeraux, 0) ;

        \path        (layer\i.east) ++ (\deltalayeraux/2,-0.5) node [above] {\blue{$\bbx_\imone$}};

        \path        (layeraux\i.west) ++ (-\deltalayeraux/2,0.1) node [above] {\red{$\bblambda_\imone$}};

    }   
    
    \path (layer\nrlayers.south) ++ (0,\deltalayer) node (output) {};
    
        \path[arrow] (layer\nrlayers.south)--(output);
        \path (output) ++ (\xlabelshift,-\ylabelshift) node [above right] 
            {$\red{\bblam_\nrlayers} = \bbPhi_\rmD(\bbz, \bbtheta_\rmD, \bbtheta_\rmP)$};

    \begin{pgfonlayer}{bg} 

        \path [] (layer1.north west)             ++ (-\deltashading,  \deltashading) coordinate (nw);
        \path [] (layeraux\nrlayers.south east)  ++ ( \deltashading, -\deltashading) coordinate (se);
        
        \path [shade] (nw) -- (nw -| se) -- (se) -- (se -|nw) -- (nw);
                
    \end{pgfonlayer}{bg}

    
    
   \path (layer0) ++ (\layerwidth, 0)  ++ (\deltalayeraux, 0) ++ (\auxlayerwidth, 0) ++ (\deltanetworks, 0) node (layer0) {};

   \path (layer0) ++ (2.0,0) node (lambdanode) {};
    \path (lambdanode) ++ (\xlabelshift-0.1,\ylabelshift-0.1) node [below right] {\red{$\bblambda_3$}};
        
    \foreach \i in {1,...,\nrlayers}
    {
        \pgfmathtruncatemacro{\imone}{\i-1} 
        
        \path (layer\imone.south) ++ (0,\deltalayer) node [primalblock] (layer\i) 
            {$\bbPhi_\rmP^{\i}\big(\, \tbx_\imone, \red{\bblambda_3}, \bbz; \bbtheta_\rmP^{\i} \, \big)$};    
        
        \path[arrow] (layer\imone.south)--(layer\i.north);
        \path        (layer\imone.south) ++ (\xlabelshift,\ylabelshift) node [below right] {$\tbx_\imone$};

        \path[arrow] (lambdanode.west) |- (layer\i.east);
    }   
    
    \path (layer\nrlayers.south) ++ (0,\deltalayer) node (output) {};
    
        \path[arrow] (layer\nrlayers.south)--(output);
        \path (output) ++ (\xlabelshift,-\ylabelshift) node [above right] 
            {$\blue{\tbx_\nrlayers} = \bbPhi_\rmP(\red{\bblambda_3}, \bbz; \bbtheta_\rmP)$};

    \begin{pgfonlayer}{bg} 

        \path [] (layer1.north west)          ++ (-\deltashading,  \deltashading) coordinate (nw);
        \path [] (layer\nrlayers.south east)  ++ ( \deltashading+0.2, -\deltashading) coordinate (se);
        
        \path [primalshade] (nw) -- (nw -| se) -- (se) -- (se -|nw) -- (nw);
                
    \end{pgfonlayer}{bg}

\end{tikzpicture}

\caption{Constrained-optimization unrolling in \eqref{eq:primal_unrolling}--\eqref{eq:recoverability}. (Left) The primal network with $K=3$ unrolled layers approximates the stationary point $\bbx_l$ for a given $\bblambda_l$. (Middle) The dual network with $L=3$ layers (red) calls the primal network (blue) at each layer, steering the iterates toward $\bblambda_3 \approx \bblambda^*$. (Right) The solution $\bbx^*$ is recovered by feeding $\bblambda_3$ into the primal network.
%
}

\label{fig:architecture}

\end{figure*}

}

Our goal is to train the primal and dual networks such that the output of the primal network satisfies $\widetilde\bbx_K \approx \bbx^\star(\bblambda)$ for any $\bblambda$, and the output of the dual network satisfies $\bblambda_L \approx \bblambda^\star$ for a given optimization problem. 
In addition, the predicted trajectories should follow descent and ascent dynamics with respect to the corresponding optimization variable. 



\section{Constrained Dual Unrolling}\label{sec:descent_constraints}

We formulate an unsupervised training problem that mirrors the dual problem in \eqref{eq:outer}--\eqref{eq:inner}. We cast the training as a nested optimization problem where the deterministic objectives are replaced with statistical counterparts evaluated over a distribution of problem instances. The outer problem corresponds to training the dual network, while the inner problem trains the primal network. Formally, we define the training problem as 
\begin{align}
     \bbtheta_\rmD^\star ~\in~ \underset{\bbtheta_\rmD}{\text{argmax}} \quad &  \mbE \Big[\ccalL \big(\, \bbPhi_\rmP (\bblambda_L, \bbz; \bbtheta_\rmP^\star)  , \, \bblambda_L; \bbz\big) \Big], \label{eq:dual_training} \\
    {\text{with} \quad \,} &
        {\bbtheta_\rmP^\star} \in \underset{\bbtheta_\rmP}{\text{argmin}} \ \mbE \Big[\ccalL \big( \, \bbPhi_\rmP (\bblambda, \bbz; \bbtheta_\rmP), \, \bblambda; \bbz \big) \Big], \label{eq:primal_training}
\end{align}
where $\bblambda_L = \bbPhi_\rmD(\bbz;\bbtheta_\rmD, \bbtheta_\rmP^\star)$ is the final output of the dual network evaluated for a problem instance $\bbz$. 

The joint objective in \eqref{eq:dual_training}--\eqref{eq:primal_training}, however, does not guarantee that the trajectories formed by the trained primal and dual networks exhibit the descent and ascent behavior of the DA algorithm. This limitation is evident in Figure~\ref{fig:comp2}~(right), where the primal (top) and dual (bottom) networks generate random trajectories with large excursions before missing the maximizer of the dual function. This is because the objectives depend solely on the final outputs of the primal and dual networks without explicitly encouraging intermediate layers to follow principled optimization dynamics. 

To address this, we impose descent and ascent constraints on the training objectives and reformulate the two levels as constrained training problems. Under these constraints, primal and dual networks, depicted in Figure~\ref{fig:comp2}~(middle), generate predominantly monotone trajectories with only mild oscillations and recover a saddle point that is substantially closer to the true optimum. 
For the rest of this section, we describe the constrained training paradigm that yields such descent and ascent trajectories.

\subsection{Constrained Dual Unrolled Networks}\label{sec:CDU}
The primal network is required to monotonically decrease the Lagrangian function across its layers. For this, we reformulate the inner problem as the constrained problem:
\begin{align}
     P_\rmP^\star = \underset{\bbtheta_\rmP}{\text{min}}\  
    & \mbE \Big[\,\ccalL \big( \, \bbPhi_\rmP (\bblambda, \bbz; \bbtheta_\rmP), \, \bblambda; \bbz \big) \,\Big], \nonumber \\
    {\text{s.t. \ }} &
        \mbE\!\Big[
            \|\! \nabla_{\bbx} \ccalL(\widetilde\bbx_{\scriptscriptstyle{k}}, \bblambda; \bbz) \!\|
            - \alpha_k \|\! \nabla_{\bbx} \ccalL(\widetilde\bbx_{\scriptscriptstyle{k-1}}, \bblambda; \bbz) \!\|
        \Big] 
            \!\leq\! 0,
            \forall k, 
            \label{eq:constrained_primal_training}
\end{align}
where $\alpha_k$ is a design parameter controlling the descent rate and the initial value $\widetilde\bbx_0$ is chosen randomly from a predetermined distribution, e.g., uniform.
In \eqref{eq:constrained_primal_training}, the objective function and constraints are obtained in expectation over the joint distribution of problem instances and Lagrangian multipliers. Alternatively, if the Lagrangian is not differentiable, one may enforce descent by requiring the Lagrangian value to decrease across layers, i.e., we force $ \ccalL(\widetilde\bbx_k, \bblambda; \bbz)  \leq \alpha_k  \ccalL(\widetilde\bbx_{k-1}, \bblambda; \bbz) $ for all $k$. While not entirely equivalent, since they may yield different trajectories, the two formulations promote convergence toward stationary points of the Lagrangian.

The dual network is required to ensure monotonic increase in the dual function across its layers. A key property of the dual function is its concavity.
Since we have $\bbf\big( \bbx^\star(\bblambda) \big) \in \partial g(\bblambda)$ by Danskin's theorem \cite{bertsekas1997nonlinear},
we impose constraints that reduce the $\ell_2$-norm of the constraint slack $\|\bbf(\bbx_l;\cdot)\|$ across the layers as a proxy for ascending the dual function. 
The training of the dual network is then defined as the constrained problem: 
\begin{align}
     P_\rmD^\star = \underset{\bbtheta_\rmD}{\text{max}} \ 
     & \mbE \Big[\, \ccalL \big(\, \bbPhi_\rmP (\bblambda_L, \bbz; \bbtheta_\rmP^\star)  , \, \bblambda_L; \bbz\big) \,\Big], 
     \nonumber \\
    {\text{s.t. \quad}} &
        \mbE\Big[\big\|[ 
                \bbf(\bbx_{\scriptscriptstyle{l}}; \bbz) ]_+\big\|
                - \beta_l  \big\|[\bbf(\bbx_{\scriptscriptstyle{l-1}} ; \bbz) 
        ]_+\big\|\Big] \leq 0, \
    \forall l, \label{eq:constrained_dual_training}
\end{align}
where $\beta_l \in (0, 1]$ is a design parameter, $\bblambda_0$ is initialized randomly, $[\bbx]_+ \!=\! \max\{0, \bbx\}$, and the expectations are over a distribution of problem instances. The quantities $\bbx_l$, for all $l$, are naturally produced during the forward pass of the dual network, and therefore no additional computation is required to evaluate the constraints.

Following \cite{hadou2024Robust}, we add Gaussian noise $\bbxi_k \sim \ccalN(\mathbf{0}, \sigma_k^2 \bbI)$ and $\bbxi_l \sim \ccalN(\mathbf{0}, \gamma_l^2 \bbI)$ to the outputs of each unrolled primal and dual layer, respectively, during training only. The noise variance decreases across layers to ensure that the final estimates are not contaminated by excessive randomness. As discussed in that work, the additive noise encourages each layer to take steps that are less dependent on the precise path taken by earlier layers. It also promotes exploration of the optimization landscape, which, when combined with descent constraints, improves the generalizability of unrolled models.

In principle, the primal network can be trained separately over the joint problem-multiplier distribution. However, this distribution is generally unknown \emph{a priori} as it captures the range of multipliers that the dual network may generate during execution. The nested formulation in \eqref{eq:dual_training}--\eqref{eq:primal_training} circumvents this challenge by enabling an alternating training scheme. First, the dual network is updated on the outer-level objective, producing multiplier trajectories across its layers. 
These multipliers are then used as training data for the primal network in the inner problem. The updated primal network subsequently generates new primal solutions that inform the next dual update. 
Algorithm~\ref{alg:joint} summarizes this alternating scheme. For each problem, we perform a fixed number of stochastic gradient steps before switching to the other, allowing both networks to jointly converge. This process implicitly samples from the true multiplier distribution, ensuring that each network is optimized against data generated by the current state of the other. In the following, we elaborate on these training procedures.

\vspace{-20pt}
\subsection{Primal Network Training}
At the $(m+1)$-th training iteration, we first consider the empirical training problem of the primal network, defined as
\begin{align}
     \bbtheta_\rmP^{\scriptscriptstyle{(m+1)}}&  \in \underset{\bbtheta_\rmP}{\text{min}} \  \widehat\mbE\Big[\, \ccalL \big( \, \bbPhi_\rmP (\bblambda^{\scriptscriptstyle{(m)}}, \bbz; \bbtheta_\rmP), \, \bblambda^{\scriptscriptstyle{(m)}}; \bbz \big)\,\Big], \nonumber \\
    {\text{s.t. }} &
        \widehat\mbE\Big[ \|\nabla\ccalL\big(\widetilde\bbx_{\scriptscriptstyle{k}}, \bblambda^{\scriptscriptstyle{(m)}}; \bbz\big)\| \! -\! \alpha_k \|\nabla\ccalL\big(\widetilde\bbx_{\scriptscriptstyle{k-1}}, \bblambda^{\scriptscriptstyle{(m)}}; \bbz\big)\| \Big] \!\leq\! 0,
     \label{eq:empirical_primal_training}
\end{align}
where the superscript ${(m)}$ denotes the iteration index and $\widehat\mbE$ represents the empirical average. Equation \eqref{eq:empirical_primal_training} is equivalent to \eqref{eq:constrained_primal_training} except that the statistical average is replaced with its empirical counterpart evaluated over $N$ problem instances, each paired with $M$ realizations of the multipliers. 
To construct this dataset, for each problem instance $\bbz$, the dual network $\bbtheta_\rmD^{\scriptscriptstyle{(m)}}$ is executed with different random initializations $\bblambda_0$, and then a set of $M$ Lagrangian multipliers is sampled from the outputs of the dual layers. These multipliers are detached from the computation graph and we do not backpropagate through them.

Problem \eqref{eq:empirical_primal_training} is a constrained optimization problem itself, which we solve through its dual problem \cite{chamon2022constrained}. We first construct the Lagrangian function using dual multipliers $\bbmu \in \reals^{K}_+$, which we refer to as \emph{meta} Lagrangian and \emph{meta} multipliers to distinguish them from those associated with the original problem in \eqref{eq:problem}. The meta Lagrangian function is defined as 
\begin{align}
    \ccalL_\rmP & \big( \bbtheta_\rmP, \ \bbmu \big) ~=~ \widehat\mbE \Big[ \, 
    \ccalL \big( \, \bbPhi_\rmP (\bblambda^{\scriptscriptstyle{(m)}}, \bbz; \bbtheta_\rmP), \, \bblambda^{\scriptscriptstyle{(m)}}; \bbz \big) \, \Big]& \nonumber \\
     & + \sum_{k=1}^K \mu_k \widehat\mbE \Big[\|\nabla \ccalL\big(\widetilde\bbx_{\scriptscriptstyle{k}}, \bblambda^{\scriptscriptstyle{(m)}}; \bbz\big)\| \!-\! \alpha_k \| \nabla \ccalL\big(\widetilde\bbx_{\scriptscriptstyle{k-1}}, \bblambda^{\scriptscriptstyle{(m)}}; \bbz\big) \| \Big]\!,
\end{align}
where $\mu_k$ is the meta multiplier associated with the $k$-th constraint. The corresponding dual problem is then given by 
\begin{align}
    D_\rmP ~=~ \max_{\bbmu\in \reals^K_+} \ \min_{\bbtheta_\rmP} \ \ccalL_\rmP \big(\bbtheta_\rmP, \bbmu \big), \label{eq:primal_maxmin}
\end{align}
which we solve with a primal-dual algorithm, outlined in Algorithm \ref{alg:primal-training}. The algorithm executes a stochastic primal-dual approach, where it constructs batches of size $N_\rmB M$. Then, it alternates between a stochastic gradient-descent step that minimizes $\ccalL_\rmP$ with respect to $\bbtheta_\rmP$ and a gradient-ascent step that maximizes it with respect to $\bbmu$.

\begin{algorithm}[t]
\caption{Joint Training (calls \cref{alg:primal-training,alg:dual-training})}
\label{alg:joint}
\begin{algorithmic}[1]
\Require initial params $\bbtheta_\rmP,\bbtheta_\rmD$, initial meta multipliers $\bbmu, \bbnu$, training rates $\epsilon_\rmP, \eta_\rmP, \epsilon_\rmD, \eta_\rmD$, dataset $\ccalD_\bbz$
\For{each iteration}
  \State $\bbtheta_\rmP, \bbmu \gets \Call{TrainPrimal}{\bbtheta_\rmP,\bbtheta_\rmD, \bbmu, \epsilon_\rmP, \eta_\rmP,\ccalD_\bbz}$ 
  \State $\bbtheta_\rmD, \bbnu \gets \Call{TrainDual}{\bbtheta_\rmP,\bbtheta_\rmD, \bbnu, \epsilon_\rmD, \eta_\rmD,\ccalD_\bbz}$ 
\EndFor
\State \Return $\bbtheta_\rmP,\bbtheta_\rmD$
\end{algorithmic}
\end{algorithm}

\begin{algorithm*}[t]
\caption{\textproc{TrainPrimal}: Primal Network Training}
\label{alg:primal-training}
\begin{algorithmic}[1]
\Require $\bbtheta_\rmP, \bbtheta_\rmD, \bbmu, \epsilon_\rmP, \eta_\rmP, \ccalD_\bbz$
\For {each epoch}
    \For{each primal batch}
        \State Sample a batch of $N_\rmB$ problem instances, $\{\bbz_{(j)}\}_{j=1}^{N_\rmB} \sim \ccalD_\bbz$ 
        %
        \State Sample $M$ multipliers per problem, $\{\bblambda_{(i,j)}\}_{i=1, j=1}^{M,N_\rmB}$, from the trajectories generated by $\bbtheta_\rmD$
        \Comment{Eq. \eqref{eq:dual_layer_unrolling}}
        \State Execute the primal network to generate primal trajectories $\{ \widetilde\bbx_{k,(i,j)} \}_{k,i,j}$ 
        \Comment{Eq. \eqref{eq:primal_unrolling}}
        \State Compute the empirical meta Lagrangian function, 
        
        $\quad \ccalL_\rmP(\bbtheta_\rmP, \bbmu) 
        \!\leftarrow\! 
        \frac{1}{N_\rmB M} \sum_{j=1,i=1}^{N_\rmB, M} \ccalL (\widetilde\bbx_{K,(i,j)}, \bblambda_{(i,j)}; \bbz )
        \!+\! \sum_{k=1}^{K}\mu_k \Big(\!\|\!\nabla
        \ccalL(\widetilde\bbx_{k,(i,j)}, \bblambda_{(i,j)};\bbz)\!\| \!-\! \alpha_k \|\!\nabla \ccalL(\widetilde\bbx_{k-1, (i,j)}, \bblambda_{(i,j)}; \bbz)\!\|\! \Big)$ 
        %
        %
        \State $\bbtheta_\rmP \leftarrow \bbtheta_\rmP - \epsilon_\rmP \cdot  \nabla_{\bbtheta_\rmP} \ccalL_\rmP(\bbtheta_\rmP, \bbmu)$ \Comment{update the primal parametrization}
        \State $\bbmu \leftarrow \big[\, \bbmu + \eta_\rmP \cdot \nabla_{\bbmu} \ccalL_\rmP(\bbtheta_\rmP, \bbmu) \,\big]_+$ 
        \Comment{update the meta Lagrangian variable}
    \EndFor
\EndFor
\State \Return $\bbtheta_\rmP, \bbmu$
\end{algorithmic}
\end{algorithm*}

\begin{algorithm*}[t]
\caption{\textproc{TrainDual}: Dual Network Training}
\label{alg:dual-training}
\begin{algorithmic}[1]
\Require $\bbtheta_\rmP,\bbtheta_\rmD, \bbnu, \epsilon_\rmD, \eta_\rmD, \ccalD_\bbz$
\For {each epoch}
    \For {each dual batch}
        \State Sample a batch of $N_\rmB$ problem instances $\{ \bbz_{(i)}\}_{i=1}^{N_\rmB} \sim \ccalD_{\bbz}$
        %
        \State Execute the dual network to generate $\{( \bblambda_{l, (i)},
        \bbx_{l,(i)}
        ) \}_{l,i}$, where $\bbx_{l,(i)} = \bbPhi_\rmP (\bblambda_{l, (i)}, \bbz; \bbtheta_\rmP)$
        %
        \State Compute the empirical meta Lagrangian function, 
        
        $\quad \ccalL(\bbtheta_\rmD, \bbnu) \leftarrow \frac{-1}{N_\rmB} \sum_{i=1}^{N_\rmB} \ccalL \big(\bbx_{L,(i)} , \bblambda_{L,(i)}; \bbz_{(i)}\big)
        + \sum_{l=1}^{L}\nu_l \cdot
        \Big(\, \big\| \big[\bbf(\bbx_{l, (i)}; \bbz_{(i)}) \big]_+\big\| 
        - \beta_l  
        \big\|
        \big[\bbf(\bbx_{l-1,(i)} ; \bbz_{(i)}) \big]_+\big\| \,\Big)$
        %
        %
        \State $\bbtheta_\rmD \leftarrow \bbtheta_\rmD - \epsilon_\rmD \cdot  \nabla_{\bbtheta_\rmD} \ccalL(\bbtheta_\rmD, \bbnu)$
        \Comment{update the dual parametrization}
        \State $\bbnu \leftarrow \big[\, \bbnu + \eta_\rmD \cdot \nabla_{\bbnu} \ccalL(\bbtheta_\rmD, \bbnu) \,\big]_+$  
        \Comment{update the meta Lagrangian variable}
    \EndFor
\EndFor
\State \Return $\bbtheta_\rmD, \bbnu$
\end{algorithmic}
\end{algorithm*}

The inner problem in \eqref{eq:primal_maxmin} can be viewed as a regularized problem, where the Lagrangian loss at each layer is regularized by the weight learned by the outer problem. The outer problem determines how much each layer contributes to minimizing the Lagrangian, thereby ensuring that all layers, not only the final one, actively drive the trajectory toward a stationary point and sustain monotonic descent dynamics.

\vspace{-10pt}
\subsection{Dual Network Training}
Given the recent primal network $\bbtheta_\rmP^{\scriptscriptstyle{(m+1)}}$, we update the dual network through the empirical constrained problem:
\begin{align}
    \bbtheta_\rmD^{\scriptscriptstyle{(m+1)}} \in \underset{\bbtheta_\rmD}{\text{argmax}} \ &  \widehat\mbE \Big[ \, \ccalL \big( \, \bbPhi_\rmP \big(\bblambda_L, \bbz; \bbtheta_\rmP^{\scriptscriptstyle{(m+1)}}\big) , \, \bblambda_L; \bbz\big) \, \Big], \nonumber \\
    s.t \quad &
     \widehat\mbE\Big[
             \big\| [\bbf(\bbx_{\scriptscriptstyle{l}}; \bbz) ]_+\big\|
             - \beta_l  \big\|[\bbf(\bbx_{\scriptscriptstyle{l-1}} ; \bbz) ]_+\big\|
        \Big] \leq 0, \label{eq:empirical_dual_training}
\end{align}
where $\widehat\mbE$ denotes the empirical average over $N$ problem instances and we let $\bbx_l = \bbPhi_\rmP (\bblambda_l, \bbz; \bbtheta_\rmP^{\scriptscriptstyle{(m+1)}})$. 
By convention, we convert \eqref{eq:empirical_dual_training} into a minimization problem by negating the objective function and write the meta Lagrangian function as
\begin{align}
    \ccalL_\rmD \big( & \bbtheta_\rmD, \bbnu \big) ~=~  - \widehat\mbE \Big[ \, 
    \ccalL \big( \, \bbPhi_\rmP \big(\bblambda_L, \bbz; \bbtheta_\rmP^{\scriptscriptstyle{(m+1)}}\big) , \, \bblambda_L; \bbz\big) \, \Big]& \nonumber \\
     & + \sum_{l=1}^L \nu_l \cdot 
     \widehat\mbE \Big[ \,
                        \big\| [\bbf(\bbx_{\scriptscriptstyle{l}}; \bbz) ]_+\big\| 
                        - \beta_l  \big\| [\bbf(\bbx_{\scriptscriptstyle{l-1}} ; \bbz) ]_+\big\|\, 
                \Big],
\end{align}
where $\bbnu \in \reals^{L}_+$ is a vector collecting the meta multipliers associated with the constraints.
The dual problem is given by
\begin{align}
D_\rmD ~=~ \max_{\bbnu \in \reals^L_+} \ \min_{\bbtheta_\rmD} \ \ccalL_\rmD \big(\bbtheta_\rmD, \bbnu \big), \label{eq:dual_maxmin}
\end{align}
which we solve by alternating between minimization over $\bbtheta_\rmD$ and maximization over $\bbnu$, as detailed in Algorithm \ref{alg:dual-training}. 

It may seem circular that we replace the original convex program in \eqref{eq:problem} with the nonconvex formulations in \eqref{eq:empirical_primal_training} and \eqref{eq:empirical_dual_training}, and then solve the latter with the same dual method we unroll. The rationale is computational; in many applications, \eqref{eq:problem} must be solved repeatedly for varying instances, each requiring a full run of an iterative solver. Our approach shifts this cost to an offline training process, where the original solver is used only to optimize the parameterizations. Once trained, solving \eqref{eq:problem} for a new instance reduces to a single forward pass, thereby achieving substantial reductions in online computational cost.

\vspace{-8pt}
\section{Theoretical Analysis}\label{sec:theory}
In this section, we establish convergence guarantees for the coupled primal--dual unrolled networks. Two challenges arise in establishing the convergence. First, 
constrained learning theory~\cite{chamon2022constrained} shows that, under suitable conditions, a learned parameterization is both near-optimal and near-feasible, and therefore, each of our descent constraints is satisfied up to a bounded error with high probability. Convergence, however, requires all layerwise constraints to hold simultaneously with high probability, which is not directly guaranteed by this theory and therefore requires a separate analysis.

Second, the objective and layerwise constraints in the dual training problem are evaluated at the primal-network prediction $\bbx_l=\Phi_{\rm P}(\bblambda_l,\bbz;\bm{\theta}_{\rm P})$, rather than the exact Lagrangian minimizer $\bbx^\star(\bblambda_l)$. The near-optimality of the dual training problem therefore does not directly establish near-optimality with respect to the true dual function. We thus need to quantify how the primal approximation error perturbs both the dual objective and its constraints. This error includes the parameterization error and the mismatch between the multiplier distribution used to train the primal network and the multipliers generated by the dual network during execution.

Our analysis therefore first establishes convergence under probabilistic layerwise descent, then quantifies the primal approximation error, including the effect of multiplier-distribution mismatch, and finally propagates this error to the feasibility and optimality guarantees of the dual network.


\subsection{Assumptions}
Our analysis rests on the following assumptions.
\begin{assumption}[Regularity]\label{as:regularity}
$f_0(\cdot\,;\bbz)$ and $\bbf(\cdot\,;\bbz)$ are continuously differentiable, $\bbf(\cdot\,;\bbz)$ is $L_f$-Lipschitz, and over the domain of interest, $\|\nabla_{\bbx}\mathcal{L}(\bbx,\bblambda;\bbz)\| \le M_{\mathrm{P}}$, and $\|\bbf(\bbx;\bbz)\| \le M_{\mathrm{D}}$.
\end{assumption}
 
\begin{assumption}[Strong convexity]
\label{as:strong-convexity}
$f_0(\cdot\,;\bbz)$ is $m$-strongly convex, $\bbf(\cdot\,;\bbz)$ is convex, strong duality holds (e.g., under Slater's condition), and the dual iterates and optimum lie in a compact set $\Lambda\subset\mathbb{R}^m_+$ of diameter $D_\Lambda$.
\end{assumption}

\begin{assumption}[Constrained-learning conditions]
\label{as:clt}
The conditions of constrained learning theory~\cite{chamon2022constrained}
hold for the primal and dual training problems. In particular, the
corresponding functional problems are strictly feasible, the parameterized
hypothesis classes approximate their functional counterparts with bounded
error, and the empirical objectives and constraints concentrate uniformly
around their population values.
\end{assumption}
Assumption~\ref{as:clt} states the conditions under which the constrained-learning
results of~\cite{chamon2022constrained} apply to our training problems. We next
record the two consequences used in the analysis: near-optimality with respect
to the functional objective and near-feasibility of the learned
layerwise constraints.

\begin{lemma}[Constrained-learning guarantees]
\label{lem:constrained_learning}
Under Assumption~\ref{as:clt}, let
$P_{D,\mathrm{fun}}^\star$ denote the optimum of the population functional
dual-training problem over the corresponding functional hypothesis class.
Then the trained dual network satisfies
\begin{equation}
P_{D,\mathrm{fun}}^\star
-
\mathbb E\!\left[
\mathcal L(\mathbf x_L,\boldsymbol\lambda_L;\mathbf z)
\right]
\le
\epsilon_D ,
\label{eq:functional_training_gap}
\end{equation}
where $\epsilon_D$ collects the approximation and finite-sample errors. Moreover, the learned primal and dual layerwise constraints are satisfied up
to errors $\zeta_P$ and $\zeta_D$, respectively, with probabilities
$1-\delta_P$ and $1-\delta_D$.
\end{lemma}
\begin{proof}
    Apply the functional strong-duality, approximation-gap, and empirical-gap arguments of~\cite[Propositions III.2--III.4]{chamon2022constrained} to the negated dual-training objective. Combining the parameterization and finite-sample errors and reverting to our maximization convention yields the result. The corresponding near-feasibility guarantees follow from the constraint
    generalization result in~\cite[Proposition~III.4]{chamon2022constrained}.
\end{proof}
Lemma~\ref{lem:constrained_learning} shows that the trained dual network attains nearly the best objective
value achievable by the corresponding functional problem. The gap
$\epsilon_D$ captures the loss due to restricting this functional problem to
the chosen network class and training it from finitely many samples.
The lemma  also guarantees approximate satisfaction of the learned layerwise constraints.
The gaps $\zeta_P$ and $\zeta_D$ quantify the residual violations of the primal
and dual training constraints, respectively. 
 
\vspace{-8pt}
\subsection{Stochastic Descent Unrolled Networks}

A direct deterministic analysis would require all layerwise descent constraints to hold simultaneously. Under Lemma~\ref{lem:constrained_learning}, however, each constraint is guaranteed only with probability \(1-\delta\), so the probability that an entire \(K\)-layer trajectory satisfies every constraint can decay as \((1-\delta)^K\). This quickly becomes uninformative as the network depth grows. We therefore model the unrolled trajectory as a stochastic descent process, allowing individual layers to violate their contraction constraints while analyzing the effect of these violations on convergence.

To formalize this argument, consider a nonnegative residual $V_k$ measuring the optimality at layer $k$, and let $\mathcal{F}_k$ denote the filtration generated by the initialization and the randomness up to primal layer $k$. The following result shows that probabilistic satisfaction of the layerwise contractions is sufficient to drive this residual toward a bounded neighborhood.

\begin{lemma}\label{lem:generic}
Let $\{V_k\}_{k\ge 0}$ be a sequence of nonnegative random variables adapted to $\{\mathcal{F}_l\}$ with $V_k \le {M}$ almost surely.
Suppose that, for every $k$ and some $\beta\in(0,1)$, the contraction
\begin{align}\label{eq:generic-constraint}
    \mbE\big[ V_k - \beta\, V_{k-1} \big] \le \zeta
\end{align}
holds with probability $1-\delta$ each. Then,
\begin{align} \label{eq:generic-liminf}
    &\lim_{k\to\infty} \mbE\Big[\min_{j\le k} V_j\Big]
    ~\le~ \eta
    := \frac{1}{1-\beta}\left(\zeta +
    \frac{\delta {M}}{1-\delta}\right)
    \quad a.s., 
\end{align}
and, for every $K>1$,
\begin{align}\label{eq:generic-lastlayer}
    \mbE\big[V_K\big] \;\le\; (1-\delta)^K \beta^K\,
    \mbE\big[V_0\big] \;+\; \eta.
\end{align}
\end{lemma}

The proof follows the stochastic descent analysis of~\cite{hadou2024Robust} and is provided in Appendix~\ref{app:generic}. 
Lemma~\ref{lem:generic} does not require the layerwise constraints to hold jointly.
Although the probability that every layer contracts may vanish with depth, the effect of constraint violations does not accumulate along the trajectory. The reason is that contractions at subsequent layers geometrically attenuate errors introduced at earlier ones, so their cumulative contribution remains bounded. Thus, the residual still converges to a neighborhood of the optimum rather than drifting away with increasing depth. The size of this neighborhood reflects both the near-feasibility error $\zeta$ and the violation probability through the term $\delta M/(1-\delta)$.

\vspace{-4pt}

\subsection{Primal Network Convergence}
We first apply Lemma~\ref{lem:generic} to the primal network with the
residual $V_k = \|\nabla_{\bbx}
\mathcal{L}(\widetilde{\bbx}_k,\bblambda;\bbz)\|$, whose
contraction at rate $\alpha_k$ is exactly the descent constraints imposed in~\eqref{eq:empirical_primal_training}. Lemma~\ref{lem:generic} then allows us to establish convergence without requiring these constraints to hold simultaneously across all primal layers.

\begin{theorem}[Primal convergence]\label{thm:primal}
Let $\bm{\theta}_{\mathrm{P}}^\star$ be trained according to~\eqref{eq:empirical_primal_training} under Assumptions~\ref{as:regularity} and \ref{as:clt}, and $\alpha_k = \alpha$ for all $k$. Then, in expectation over the joint distribution of problem instances and training multipliers $(\bbz,\bblambda) \sim
\ccalD$, the output of the $K$-layer primal network satisfies
\begin{align}\label{eq:primal-lastlayer}
    \mbE
    \big[
        \left\|
            \nabla \ccalL(\widetilde\bbx_K,\bblambda 
            \right.&
            \left. ;\bbz)
        \right\|
    \big]
    ~\leq~
    (1-\delta_P)^K
    \alpha^K \, 
    \mbE\big[V_0\big]
    +
    \eta_P,\\
    \text{where} \quad \eta_P
    & ~:=~
    \frac{1}{1- \alpha}
    \left(
        \zeta_P
        +
        \frac{\delta_P M_P}{1-\delta_P}
    \right).
\end{align}
Moreover, when the network is executed on multipliers drawn from
$\ccalD'$ that differs from the training distribution $\ccalD$ the same bound
holds with $\zeta_{\mathrm{P}}$ replaced by $\zeta_{\mathrm{P}}
+ 2 M_{\mathrm{P}}\, d(\ccalD,\ccalD')$.
\end{theorem}
The proof is relegated to Appendix~\ref{app:primal}. Theorem~\ref{thm:primal} shows that the primal residual consists of a transient term that decays geometrically with the network depth and a nonvanishing error floor $\eta_{\mathrm P}$. The latter captures the imperfect satisfaction of the descent constraints through both the near-feasibility error $\zeta_{\mathrm P}$ and the probability $\delta_{\mathrm P}$ of violating an individual constraint. 

The bound above is stated for multipliers drawn from the distribution used to train the primal network. During execution of the coupled architecture, however, the primal network is queried at the multipliers generated by the dual network. These multipliers may follow a different distribution. The second part of Theorem~\ref{thm:primal} accounts for this shift through the distance between the two distributions
$d(\mathcal{D},\mathcal{D}')$. Hence, the quality of the primal approximation depends not only on the depth of the network and the accuracy of its learned descent constraints, but also on how closely the training multiplier distribution matches the one encountered during execution.


\begin{lemma}[Primal approximation errors]\label{lem:bias}
Under Assumptions~\ref{as:regularity}--\ref{as:clt}, the primal estimates
$\bbx_l = \bbPhi_{\mathrm{P}}(\bblambda_l,\bbz;
\bm{\theta}_{\mathrm{P}}^\star)$ generated during the forward pass of the dual
network satisfy, for all $l$,
\begin{align}\label{eq:bias}
    \mbE  \Big[
                \ccalL( & \bbx_l, \bblambda_l ;\bbz) 
                - g(\bblambda_l;\bbz)
            \Big]
    ~\le~ b_g := \frac{M_\mathrm{P}}{2m}
    \rho_K, \\
    \mbE \Big[
    \big\| [\bbf(\bbx_l 
    \Big. & \Big. 
    ;\bbz)]_+
    -
    [\bbf (\bbx^\star(\bblambda_l);
    \bbz)]_+\big\|
    \Big]
    \leq
    b_f
    :=
    \frac{L_f}{m}\rho_K,
    \label{eq:feas_bias} \\
    \text{where} \,\, \rho_K & = (1-\delta_P)^K
    \alpha^K \, 
    \mbE\big[V_0\big] \nonumber\\
    &+
    \frac{1}{1- \alpha}
    \left(
        \zeta_{\mathrm{P}} + 2M_\mathrm{P} 
        d(\ccalD, \ccalD')
        +
        \frac{\delta_\mathrm{P} M_P}{1-\delta_P}
    \right).
\end{align}
\end{lemma}

The proof is relegated to Appendix~\ref{app:bias}. Lemma~\ref{lem:bias} makes the coupling between the two networks explicit.
The dual network evaluates its objective and constraints using the output of the primal network rather than the exact Lagrangian minimizer. The bounds \(b_g\) and \(b_f\) quantify the resulting errors in the dual objective and constraint residuals, respectively. Both inherit the primal convergence error and therefore depend on three factors: the finite depth of the primal network, the near-feasibility of its learned descent constraints, and the mismatch between the multiplier distributions used during training and execution. The depth-dependent component decays geometrically with \(K\), while the remaining terms determine the limiting approximation error seen by the dual network.

The distribution-mismatch term explains the role of the alternating training procedure in Algorithm~\ref{alg:joint}. By repeatedly regenerating the primal training multipliers from the current dual network, the algorithm keeps the primal training distribution close to the multipliers encountered during execution, thereby limiting this source of primal approximation error.

\subsection{Dual Network Convergence}
We now characterize the performance of the dual network. The layerwise constraints imposed during training drive the corresponding primal solutions toward feasibility, while the final-layer training objective promotes near-optimality of the predicted multiplier. The feasibility guarantee follows from the same stochastic contraction argument used in the primal analysis.
For dual optimality, we directly combine the constrained-learning near-optimality guarantee in Lemma~\ref{lem:constrained_learning} with the primal approximation bound in Lemma~\ref{lem:bias}.

\begin{theorem}[Dual convergence]\label{thm:dual_convergence}
Let $\bbtheta_D^\star$ be trained according to~\eqref{eq:empirical_dual_training} under Assumptions~\ref{as:regularity}--\ref{as:clt}, and assume $\beta_l=\beta$ for all $l$. 
Let $\boldsymbol\xi^\star$ denote an
optimal multiplier associated with the constraints of the 
functional dual-training problem.
Then the trajectory generated by the dual network satisfies
\begin{align}\label{eq:dual_feasibility_bound}
    \mbE\Big[
        \left\|
            [\bbf(\bbx_L;\bbz)]_+
        \right\|
    \Big]
    \leq
    (1-\delta_D)^L\beta^L\mbE[V_0]+\eta_D,
\end{align}
where
$
\eta_D=
(1-\beta)^{-1}
\left(
\zeta_D+\delta_D M_D/(1-\delta_D)
\right).
$
Moreover, the final predicted multiplier satisfies
\begin{equation}
    \mathbb{E} \Big[
        g^\star(\mathbf z)
        -
        g(\boldsymbol{\lambda}_L;\mathbf z)
    \Big]
    ~\le~
    \epsilon_D
    +
    b_g
    +
    (1-\beta)\|\bbxi^\star\|_1 b_f,
    \label{eq:dual_gap}
\end{equation}
where $b_g$ and $b_f$ are given in Lemma~\ref{lem:bias}.
\end{theorem}
The proof is provided in Appendix~\ref{app:dual_convergence}.
The first bound follows from the same stochastic contraction argument used for the primal network. The constraint violation therefore decreases with the dual depth until reaching the neighborhood $\eta_D$. The second result gives dual optimality guarantee. 
The term $\epsilon_D$ captures the near-optimality error of constrained learning, including the parameterization and finite-sample effects, while $b_g$ captures the error induced by evaluating the dual objective using the approximate primal response. 
The term $(1-\beta)\|\boldsymbol\xi^\star\|_1 b_f$ arises because the primal network may
still incur a constraint residual of size $b_f$, which relaxes each functional
dual-training constraint by at most $(1-\beta)b_f$. The multiplier
$\boldsymbol\xi^\star$ converts this relaxation into the corresponding change
in the optimal functional training value.
Since the training objectives and constraints are defined in expectation, these guarantees hold on average over the problem distribution rather than for every individual instance.

\vspace{-7pt}
\section{{Quadratic Programs with Linear Constraints}}\label{sec:miqp}
We consider a quadratic program (QP) with linear inequality constraints, formulated as:
\begin{align} 
        \min_{\bbx} \quad \tfrac{1}{2} \ \bbx^\top \bbP \bbx + \bbq^\top \bbx \quad \ \
        \text{s.t.}\quad \ \  {\bbA \bbx} \leq {\bbb}, \label{eq:QP}
\end{align}
where $\bbx \in \reals^n$, $\bbP \in \reals^{n \times n}$, $\bbq \in \reals^n$, $\bbA \in \reals^{m \times n}$, $\bbb \in \reals^m$.
The objective function is convex, following the assumption that $\bbP$ is positive semi-definite. 
The Lagrangian function associated with \eqref{eq:QP} is defined as
\begin{align}
    \ccalL(\bbx, \bblambda) = \tfrac{1}{2}\bbx^\top \bbP \bbx + \bbq^\top \bbx + \bblambda^\top \big( 
                                \bbA \bbx - \bbb
                                \big), \label{eq:QP_lagrangian}
\end{align}
with $\bblambda \in \reals_+^{m}$ is the Lagrangian multiplier vector that collects the multipliers associated with each constraint. Our goal is to train a primal and a dual network that collaboratively predict the saddle point of the Lagrangian function for any instance drawn from a family of QP problems. 



\vspace{-7pt}
\subsection{GNN-based Architectures} 
We parametrize the primal and dual networks with two GNNs. To facilitate this, we represent the QP problem as a graph that captures the relationships between the optimization variables and the constraints. The graph comprises $n$ nodes corresponding to the decision variables and $m$ nodes representing the constraints. The graph adjacency is then defined as
\begin{align}
    {\bf S} ~=~ \left[
    \begin{matrix}
        {\bf P} & {\bf A}^\top \\
        {\bf A} & {\bf 0}
    \end{matrix} \right]. \label{eq:GSO}
\end{align}
For notational convenience, and without loss of generality, we index the nodes such that the $n$ variable nodes precede the $m$ constraint nodes in the adjacency matrix. 

The core layer of GNNs consists of a graph filter followed by a nonlinearity \cite{gama_graphs_2020}. The graph filter is defined as a polynomial function of a graph shift operator (GSO), typically represented by a matrix such as the adjacency matrix or the graph Laplacian. The output of the $\ell$-th graph layer is then given by
\begin{align}
\bbX_{\ell} =  \varphi  \left( \,  \sum_{h = 0}^{K_\rmh} \;  \bbS^h \bbX_{\ell-1} \bbTheta_{\ell,h}  \right)\!, \label{eq:graph_layer}
\end{align}
where $\bbTheta_{\ell,h} \in \reals^{F_{\ell-1} \times F_{\ell}}$ is the set of learnable parameters, $K_\rmh$ represents the number of hops used in the graph filter, and $\varphi$ is a nonlinear activation function. As a design convention, we treat a block of $T$ stacked graph layers as a single unrolled layer within both the primal and dual networks. The final $F_T$-dimensional output of the graph block is then processed with a linear layer to generate the new estimate.

In the primal network, each unrolled layer consists of a cascade of $T_\rmP$ graph sub-layers. To describe the data flow through these layers, we denote by $\widetilde\bbX_{\ell}^{(k)}$ the graph signal at the $\ell$-th graph sub-layer of the $k$-th unrolled layer, where the superscript indicates the unrolled layer and the subscript denotes the graph sub-layer within it. 
The input to the $k$th unrolled layer, $\widetilde\bbX^{(k)}_0 \in \reals^{(n+m) \times 2}$, is defined as
\begin{align}
    \widetilde\bbX_0^{(k)} ~=~ \left[
    \begin{matrix}
        \widetilde\bbx_{k-1} & \bbq \\
        \bblambda & \bbb
    \end{matrix} \right],
\end{align}
where $\widetilde\bbx_{k-1}$ is the output of the previous unrolled layer, and $\bblambda$, $\bbq$ and $\bbb$ are input data.
The output of the graph block, $\widetilde\bbX_{T_\rmP}^{(k)} \in \reals^{(n+m) \times F_{T_\rmP}}$, is then processed with a linear layer to predict the new estimate of the primal variable. Formally, the output of the $k$th unrolled layer is given by
\begin{equation}
    \widetilde\bbx_k ~=~ \widetilde\bbx_{k-1} +  \bbM_\rmP \, \widetilde\bbX_{T_\rmP}^{(k)} \,  \bbW_k + \bbc_k ,
\end{equation}
where $\bbW_k \in \reals^{F_{T_\rmP}}$ and $\bbc_k \in \reals^{n}$ are learnable parameters. The matrix $\bbM_\rmP \in \{0,1\}^{n \times (n+m)}$ is a selection matrix that extracts the outputs associated with the $n$ variable nodes and excludes the outputs of the constraint nodes.

Similarly, each unrolled dual layer contains a cascade of $T_\rmD$ graph sub-layers. The input to the $l$-th unrolled dual layer contains the recent estimates of the primal and dual variables $\bbx_{l-1}$ and $\bblambda_{l-1}$ along with the input parameter $\bbq$ and $\bbb$, i.e.,
\begin{align}
    \bbX_0^{(l)} ~=~ \left[
    \begin{matrix}
        \bbx_{l-1} & \bbq \\
        \bblambda_{l-1} & \bbb
    \end{matrix} \right].
\end{align}
The output of this graph block, $\bbX_{T_\rmD}^{(l)} \in \reals^{(n+m) \times F_{T_\rmD}}$, is processed with a linear layer, followed by a relu function, to generate the following estimate
\begin{equation}
    \bblambda_l ~=~ \varphi_\text{relu} \Big( \bblambda_{l-1} + \bbM_\rmD \, \bbX_{T_\rmD}^{(l)} \, \bbW_l + \bbc_l \Big),
\end{equation}
where $\bbW_l \in \reals^{F_{T_\rmD}}$ and $\bbc_l \in \reals^{m}$ are learnable parameters--different from those used in the unrolled primal layers despite the shared notation. Moreover, the selection matrix $\bbM_\rmD \in \{0,1\}^{(m) \times (n+m)}$ chooses the values corresponding to the constraint nodes and omits the variable nodes, and the relu activation function ensures that the multipliers are nonnegative.


\vspace{-7pt}
\subsection{Experiment Setup}
In our experiments, we consider a family of QP problems with $n = 1000$ decision variables, and $m=812$ linear inequality constraints. For each problem instance, we draw the elements of $\bbA$ and $\bbq$ independently from a standard normal distribution. The matrix $\bbP$ is designed to be positive definite and the matrix $\bbA$ is normalized by dividing its elements by its spectral norm. The vector $\bbb$ is then evaluated as $\bbA \bbx + \bbepsilon$, with $\bbepsilon \sim \ccalU\text{nif}[0,1]^m$, to ensure that each problem is feasible. 

We use a primal network consisting of $K = 4$ unrolled primal layers, each composed of $T_\rmP = 3$ graph sub-layers. The dual network comprises $L = 16$ dual layers with $T_\rmD = 3$ graph sub-layers. In both networks, each graph filter aggregates information from $K_\rmh = 1$-hop neighbors, processes $F_l = F_k = 64$ hidden features, and is followed by a tanh activation.

\textbf{Training.}
To train the primal network, we construct a training dataset of $200$ problem instances, divided into mini-batches of $N_\rmB=4$ problems. At each iteration, $M=32$ multipliers per problem are generated to evaluate the empirical meta Lagrangian in the training loss \eqref{eq:empirical_primal_training}.
Half of these multipliers are sampled from the trajectories generated by the recent dual network. The elements of the remaining multipliers are drawn independently from a uniform distribution, i.e., $\lambda_i \sim \mathrm{Unif}[0,1]$, with probability $0.7$, and set to zero ($\lambda_i = 0$) otherwise. Training is performed with descent constraints imposed on the gradient norm of the Lagrangian, rather than on the Lagrangian value itself.
For the dual network, we use a dataset of $400$ problem instances, grouped in batches of $N_\rmB = 32$. We alternate between the two networks for $100$ iterations by running one epoch of primal network training (Algorithm~\ref{alg:primal-training}) after every $15$ epochs of dual network training (Algorithm~\ref{alg:dual-training}). Each epoch consists of a full pass over the corresponding dataset. {The offline training time is approximately $4$ hours.}

For training, we optimize the architecture parameters of the primal and dual networks using two ADAM optimizers \cite{kingma2014adam} with learning rates: $\epsilon_\rmP = 10^{-4}$ and $\epsilon_\rmD = 7 \times 10^{-4}$. The meta Lagrangian multipliers are updated through gradient ascent with step sizes: $\eta_\rmP = 10^{-4}$ and $\eta_\rmD = 10^{-3}$. We share the constraint parameter $\alpha_k = 0.5$ and $\beta_l = 0.5$ across all the primal and dual layers, respectively.

\begin{table*}[t]
\centering
\small
\caption{\small Baseline Comparisons, all quantities are averaged over 100 test instances.
The objective gap is
$|\bbf(\bbx_L)-\bbf(\bbx^{\star})|/(1+|\bbf(\bbx^{\star})|)$; the violation is
$\max\{\mathbf{0},\bbA\bbx_L-\bbb\}/(1+\max_i|b_i|)$ by its mean, its 99th percentile and its maximum; and
the complementary-slackness is $\sum_i|\lambda_{L,i}(\bbA\bbx_L-\bbb)_i|/|\bbf(\bbx^{\star})|$ over all constraints,
over the constraints the method leaves active, and over the constraints it satisfies strictly and is reported for dual methods only. The reference
$(\bbx^\star,\bblambda^\star)$ is ADMM in double precision to $10^{-9}$. Classical solvers are stopped at our
network's forward time on one RTX 3090.}
\label{tab:accuracy}
\begin{tabular}{l r rrr rrr}
\toprule
 & & \multicolumn{3}{c}{constraint violation \%} & \multicolumn{3}{c}{complementary slackness \%} \\
\cmidrule(lr){3-5}\cmidrule(lr){6-8}
Method & obj.\ gap \% & mean & p99 & max & all & active & inactive \\
\midrule
\textbf{CDU} (ours)  & $4.28{\scriptstyle\,\pm0.61}$ & $\textbf{0.024}{\scriptstyle\,\pm0.005}$ & $\textbf{0.64}{\scriptstyle\,\pm0.11}$ & $\textbf{1.40}{\scriptstyle\,\pm0.43}$ & $4.08{\scriptstyle\,\pm0.66}$ & $\textbf{0.51}{\scriptstyle\,\pm0.20}$ & $3.57{\scriptstyle\,\pm0.53}$ \\

OSQP  & 
$\textbf{0.39}{\scriptstyle\,\pm0.31}$ & $0.059{\scriptstyle\,\pm0.021}$ & $1.43{\scriptstyle\,\pm0.43}$ & $2.49{\scriptstyle\,\pm0.78}$ & 
$\textbf{0.87}{\scriptstyle\,\pm0.29}$ & $0.66{\scriptstyle\,\pm0.30}$ & 
$\textbf{0.21}{\scriptstyle\,\pm0.08}$ \\
SCS  & 
$43.08{\scriptstyle\,\pm9.50}$ & $2.802{\scriptstyle\,\pm0.240}$ & $27.31{\scriptstyle\,\pm2.36}$ & $41.45{\scriptstyle\,\pm7.87}$ & 
--&
--&
-- \\
DA & 
$71.69{\scriptstyle\,\pm14.20}$ & $3.010{\scriptstyle\,\pm0.348}$ & $30.26{\scriptstyle\,\pm3.46}$ & $46.41{\scriptstyle\,\pm7.89}$ & $49.29{\scriptstyle\,\pm9.16}$ & $46.57{\scriptstyle\,\pm8.77}$ & $2.72{\scriptstyle\,\pm0.51}$ \\
unconstrained ablation        & $3.30{\scriptstyle\,\pm0.70}$ & $0.553{\scriptstyle\,\pm0.072}$ & $7.00{\scriptstyle\,\pm0.94}$ & $11.97{\scriptstyle\,\pm2.60}$ & $12.75{\scriptstyle\,\pm2.03}$ & $5.89{\scriptstyle\,\pm1.11}$ & $6.86{\scriptstyle\,\pm1.03}$ \\

naive GNN                     & $0.59{\scriptstyle\,\pm0.60}$ & $0.188{\scriptstyle\,\pm0.030}$ & $2.56{\scriptstyle\,\pm0.35}$ & $5.03{\scriptstyle\,\pm6.67}$ &
--&
--&
-- \\

\bottomrule
\end{tabular}
\end{table*}

\textbf{Baselines.} We compare the performance of our method with those of the following benchmarks:
\begin{enumerate}[label=\roman*.]
    \item The DA algorithm, \eqref{eq:DA_primal}--\eqref{eq:DA_dual}, with dual learning rate $\eta = 0.01$. The minimizer of the Lagrangian, $\bbx^\star_l$, is calculated analytically by setting the gradient to zero, i.e.,
    $
        \bbx^\star_l = -\bbP^{-1} \big( \, \bbq + \bbA^\top \bblambda_l \,\big)
    $.
    \item Two classical algorithms: the operator splitting quadratic program (OSQP)~\cite{stellato2020osqp}, and splitting conic solver (SCS)~\cite{o2021operator}. 
    \item A GNN that consists of $16$ graph layers, as in \eqref{eq:graph_layer}, with a relu activation function. The GSO is as in \eqref{eq:GSO}, and the input to the first layer is the vector $[\bbb; \bbq]$ that contains the constraint slack and the linear coefficient of the objective. The final graph layer is followed by a linear layer. The network is trained via a supervised paradigm to mimic the optimal solution $\bbx^\star$, obtained by a CVXPY solver. The analysis in \cite{chen2024expressive} shows that such GNN efficiently approximates the solution of \eqref{eq:QP}.
    \item Unconstrained dual unrolling, where we train the primal and dual networks without descent constraints under the same hyperparameters. This is equivalent to the approach in \cite{Park_Van_Hentenryck_2023}, except that we use unsupervised losses rather than their supervised objective.
\end{enumerate}

\begin{figure}
    \centering    
    \vspace{-10pt}
    \includegraphics[width=0.9\columnwidth]{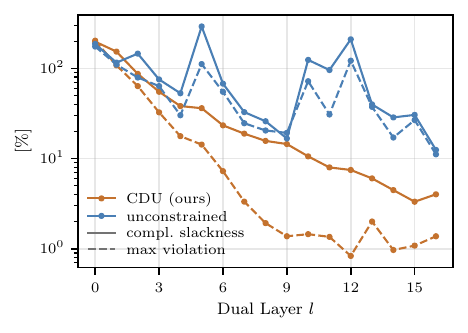}
    \vspace{-8pt}
    \caption{\small Evolution of the maximum constraint violation and complementary slackness across the unrolled dual layers. The constrained model exhibits a consistent decrease in both quantities, whereas the unconstrained model shows larger oscillations and higher residuals at the final layers.}
    \vspace{-10pt}
    \label{fig:miqp_ablation}
\end{figure}

\vspace{-7pt}
\subsection{Numerical Results}


{We evaluate the trained primal and dual networks on $100$ unseen problem instances drawn from the same distribution as the training set. Table~\ref{tab:accuracy} compares CDU with classical optimization algorithms and learning-based baselines under a common inference-time budget. 
CDU achieves a favorable balance between optimality and feasibility. Its mean constraint violation is $0.024\%$, compared with $0.059\%$ for OSQP, $2.802\%$ for SCS, and $3.010\%$ for DA. The same trend is observed in the tail of the violation distribution, where CDU attains a $0.64\%$ $99$th-percentile violation and a $1.40\%$ maximum violation. OSQP attains a smaller objective gap of $0.39\%$, compared with $4.28\%$ for CDU, but at the same runtime budget CDU produces solutions that are more feasible. SCS and DA are substantially worse in both objective value and feasibility at this computational budget. These results show that the learned solver efficiently captures the constrained solution structure while operating within a fixed inference-time budget.}

{The learning-based baselines further highlight the role of the constrained-unrolling architecture. The unconstrained ablation and naive GNN attain objective gaps of $3.30\%$ and $0.59\%$, respectively, but their mean constraint violations increase to $0.553\%$ and $0.188\%$, with substantially larger tail violations. Thus, their smaller objective gaps do not translate into better constrained solutions. In contrast, imposing the layerwise descent and ascent constraints reduces the mean violation by more than an order of magnitude relative to the unconstrained ablation. Figure~\ref{fig:miqp_ablation} provides a trajectory-level view of this effect. The constrained model progressively reduces the complementary-slackness residual across the dual layers, whereas the unconstrained model exhibits substantially more oscillatory behavior.}

\begin{table}[t]
\centering
\small
\setlength{\tabcolsep}{4pt}
\caption{\small {Mean wall-clock time (ms) to reach target accuracy.} The target is set per instance to the error our network attains. 
$m/n = 0.81$ held constant across all problems.
}
\label{tab:time-to-target}
\resizebox{\columnwidth}{!}{%
\begin{tabular}{l r r r r}
\toprule
 & $n=1000$ & $n=2000$ & $n=5000$ & {$n=10000$} \\
\midrule
\textbf{CDU}  & \textbf{34} & \textbf{37} & \textbf{144} & \textbf{498}  \\
OSQP  &  38 & 124 & 1166 & 8132  \\
SCS  & 287 & 864 & 4368 & 25357  \\
DA & 418 & 867 & $>$30s & $>$30s  \\
CPLEX  & 1349 & -- & -- & -- \\
\bottomrule
\end{tabular}
}
\vspace{-10pt}
\end{table}

{The main computational advantage of CDU becomes more pronounced as the problem size increases. Table~\ref{tab:time-to-target} reports the wall-clock time required by each solver to reach, on each problem instance, the accuracy attained by CDU. For the in-distribution problem size $n=1000$, CDU requires $34$ ms, compared with $38$ ms for OSQP, $287$ ms for SCS, $418$ ms for DA. More importantly, the gap increases rapidly with problem size. At $n=5000$ and $m=4060$, CDU reaches its target in $144$ ms, whereas OSQP requires $1166$ ms and SCS requires $4368$ ms, while DA does not reach the target within $30$ s. At $n=10000$ with $m=8120$, CDU requires $498$ ms compared with $8132$ ms for OSQP and $25357$ ms for SCS. This corresponds to approximately $16\times$ and $51\times$ lower time-to-target, respectively.}

\begin{figure*}[t!]
    \centering
    \includegraphics[width=\textwidth ]{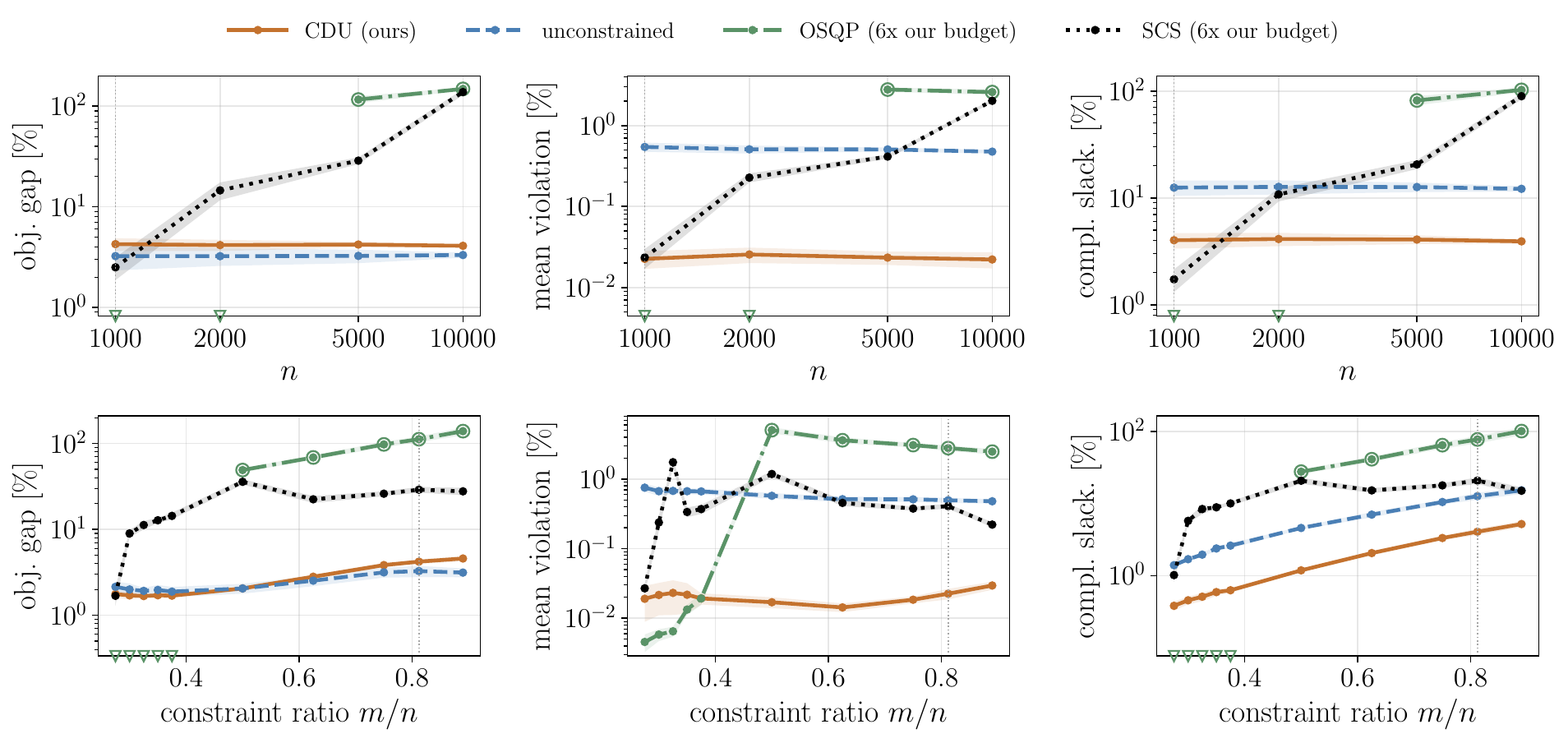}
    \vspace{-15pt}
    \caption{\small OOD generalization. 
    (Top) The objective gap, mean constraint violation, and complementary slackness as the problem size $n$ increases while the constraint-to-variable ratio is fixed at $m/n=0.81$. 
    (Bottom) The same metrics as the constraint ratio $m/n$ increases at fixed problem size $n=5000$. 
    OSQP and SCS are given $6\times$ the inference-time budget of CDU. While OSQP attains the optimum in some regimes, its performance degrades more sharply as the problem size or number of constraints increases. CDU maintains more stable performance and degrades more gracefully overall.}
    \label{fig:miqp_ood}
    \vspace{-8pt}
\end{figure*}


{
Finally, Figure~\ref{fig:miqp_ood} evaluates OOD generalization as the number of variables and constraints increase. To account for the differences in computational budgets reported in Table!\ref{tab:time-to-target}, OSQP and SCS are each given $6\times$ the inference-time budget of CDU. The top row increases the problem size $n$ while fixing $m/n=0.81$. OSQP attains essentially optimal performance in some of the smaller regimes, but its objective gap, constraint violation,
and complementary slackness deteriorate substantially as the problem size increases. SCS also degrades with increasing problem size, while CDU remains stable across the tested problem sizes. The bottom row fixes $n=5000$ and increases the constraint-to-variable ratio $m/n$. OSQP again can converge in the easier regimes, but its performance degrades more sharply as the problem becomes more heavily constrained. SCS, in contrast, remains at a relatively high error level across most of the tested regimes, with comparatively limited variation.
CDU maintains more stable performance and degrades more gracefully overall, despite operating with a substantially smaller inference-time budget.
}

\section{Power Allocation in Wireless Networks}\label{sec:pa}

Consider a network of $n$ pairs of transmitters and receivers that share the same communication channel and cause interference to each other. The interference channel is represented by a channel state $\bbH  \in \mathbb{R}^{n \times n}$ whose element $h_{ij}$ represents the channel gain between the $i$-th transmitter and the $j$-th receiver. Under a channel state $\bbH$, the $i$-th pair transmits with power $p_i$ and achieves a rate of
\begin{equation}\label{eq:rates_definition}
    r_i(\bbp; \bbH) = \log \left(1+ \frac{|h_{ii}|^2p_i}{W N_0 + \sum_{j\neq i}|h_{ji}|^2p_j} \right),
\end{equation}
where $W$ is the bandwidth, $N_0$ is the noise power spectral density, $h_{ji}$ is the $ji$-th element of $\bbH$, and $\bbp \in \mathbb{R}^n$ is a vector of transmit powers. The rate of the $i$-th user is determined by its transmit power and the interference caused by its neighboring users through their channel gains $h_{ji}$ and transmit powers $p_j$.

We aim to find a policy $\bbp$ that maximizes the sum rate of the network while forcing a subset of users $\ccalI$ to maintain a rate of at least $r_{\text{min}}$. Formally, the resource allocation problem is formulated as 
\begin{align}\label{eq:resource_allocation}
        P^\star ~=~ \underset{{\bbp}}{\text{max}} \quad & \boldsymbol{1}^\top \bbr\big(\bbp; \, \bbH\big) & \\
    {s.t. \quad} & r_i\big(\bbp; \, \bbH\big) ~\geq~ r_{\text{min}}, & \forall i \in \ccalI,  \label{eq:constrained_agents} \\
    & r_i\big(\bbp; \,\bbH\big) ~\geq~ 0, & \forall i \notin \ccalI, \label{eq:unconstrained_agents} \\
    & \boldsymbol{0} ~\leq~ \bbp ~\leq~ P_{\text{max}} \cdot \boldsymbol{1} \label{eq:power_constraints}
\end{align}
where $\boldsymbol{1} \in \mathbb{R}^n$ is the all-one vector, and the set $\ccalI \subseteq \{1, \dots, n\}$ contains the indices of the users subject to the minimum rate constraints \eqref{eq:constrained_agents}. The remaining users are expected to maximize their rates, subject only to the requirement that they do not prevent the constrained users from meeting their minimum rate guarantees. Accordingly, the constraints associated with these users \eqref{eq:unconstrained_agents} merely enforce non-negativity, which is trivially satisfied since the rate expression in \eqref{eq:rates_definition} is inherently non-negative.
Lastly, the box constraints in \eqref{eq:power_constraints} define the power budget $P_\text{max}$ for each user and can be implicitly enforced by applying a sigmoid activation function at the final layer of the predictive model, i.e., the primal network. Hence, we omit this constraint in the subsequent analysis.

To define the dual problem associated with \eqref{eq:resource_allocation}--\eqref{eq:power_constraints}, we first convert the problem to a minimization and construct a Lagrangian multiplier vector $\bblambda \in \reals^n_+$ that  collects the multipliers associated with the rate constraints in \eqref{eq:constrained_agents} and \eqref{eq:unconstrained_agents}.
The corresponding Lagrangian function is given by
\begin{align}
    \ccalL\big(\bbp, & \,\bblambda; \, \bbH\big) ~=~ -\boldsymbol{1}^\top \bbr\big(\bbp; \, \bbH\big) \nonumber \\ 
                                    &    + \sum_{i\in \ccalI} \lambda_i \Big(r_{\text{min}} - r_i\big(\bbp; \, \bbH\big) \Big)
                                      - \sum_{i\notin \ccalI} \bar\lambda_i r_i\big(\bbp; \,\bbH\big), \label{eq:power_lagrangian}
\end{align}
where we distinguish the multipliers associated with unconstrained users using a bar notation (i.e., $\bar\lambda_i$ for $i \notin \ccalI$) to differentiate them from those associated with constrained users (i.e., $\lambda_i$ for $i \in \ccalI$). Since the constraints in \eqref{eq:unconstrained_agents} are trivially satisfied---as rates are nonnegative by definition---the optimal multipliers $\bar\lambda^\star_i$ for unconstrained agents are trivially zero. For notational convenience, however, we keep these constraints in our formulation to ensure that the dimensionality of the graph signal $\bblambda$ matches that of the power signal $\bbp$.

Our goal is to train a primal and a dual network to predict $\bbp^\star$ and $\bblambda^\star$ given a channel state matrix $\bbH$, drawn from a family of networks. For convenience and without loss of generality, we arrange $\bbH$ so that the rows and columns corresponding to the constrained pairs precede those of the unconstrained pairs.  

\vspace{-7pt}
\subsection{GNN-based Architectures}
We deploy two GNNs as the underlying architectures for the primal and dual networks. Similar to \eqref{eq:graph_layer}, each graph layer consists of a graph filter followed by a nonlinearity. The output of the $\ell$-th layer, $\bbX_{\ell} \in \reals^{n \times F_{\ell}}$, is then given by
\begin{align} 
\bbX_{\ell} =  \varphi  \left( \,  \sum_{h = 0}^{K_\rmh} \;  \bbH^h \bbX_{\ell-1} \bbTheta_{\ell,h}  \right)\!, \label{eq:graph_layer2}
\end{align}
where the channel state matrix $\bbH$ is used as a GSO, $\bbTheta_{\ell,h} \in \reals^{F_{\ell-1} \times F_{\ell}}$ is the set of learnable parameters, $K_\rmh$ represents the graph filter taps, and $\varphi$ is a nonlinear activation function. Each graph layer thus maps an $F_{\ell-1}$-dimensional graph signal to an $F_\ell$-dimensional graph signal. Following the same design convention used in the QP experiment, we treat a block of stacked graph layers as a single unrolled layer within both the primal and dual networks.
GNNs are particularly well-suited for this setting, as the network state $\bbH$ defines a natural graph structure, while the power allocation and Lagrangian multiplier vectors can be interpreted as graph-structured signals.

In our experiments, a primal layer consists of a cascade of $T_\rmP$ graph sub-layers of the form in \eqref{eq:graph_layer2}. The input to the $k$-th primal layer, $\widetilde\bbX^{(k)}_{0} \in \reals^{n \times 3}$, stacks the recent prediction $\widetilde\bbp_{k-1} \in \reals^n$, the given multiplier $\bblambda \in \reals^n_+$ and the constraint slack vector $\bbs \in \reals^n$. The slack variable $s_i$ is $r_\text{min}$ when $i \in \ccalI$ and $0$ otherwise. The output of the graph block, $\widetilde\bbX^{(k)}_{T_{\rmP}} \in \reals^{n \times F_{T_\rmP}}$, is processed by a nonlinear layer with parameters $\bbW_k \in \reals^{F_{T_\rmP}}$ and $\bbc_k \in \reals^n$. The final output of the $k$-th unrolled primal layer is then given as
\begin{align}
      \widetilde\bbp_k ~=~ P_\text{max} \cdot \varphi_\text{sig} \Big(\,
      \widetilde\bbp_{k-1} + 
      \, \widetilde\bbX^{(k)}_{T_{\rmP}} \, \bbW_k + \bbc_k
      \,\Big), \label{eq:power_primal_layer}
\end{align}
where $\varphi_\text{sig}$ the sigmoid function. The deployment of a sigmoid function ensures that $\widetilde\bbp_k$ satisfies the power constraints in \eqref{eq:power_constraints}.

Similarly, each dual layer contains $T_\rmD$ cascaded graph sub-layers, followed by a nonlinear layer. The input to the $l$-th dual block, $\bbX_0^{(l)} \in \reals^{n \times 3}$, is constructed by stacking the recent predictions $\bbp_{l-1}$ and $\bblambda_{l-1}$ and the slack vector. The output of the $l$-th layer, $\bbX_{T_{\rmD}}^{(l)} \in \reals^{n \times F_{T_{\rmD}}}$, is expressed as
\begin{align}
      \bblambda_l ~=~  \varphi_\text{relu} \Big(\,
      \bblambda_{l-1} + 
      \bbm \odot  \left(\bbX^{(l)}_{T_{\rmD}} \, \bbW_l + \bbc_l \right)
      \Big), \label{eq:power_dual_layer}
\end{align}
where $\bbW_l \in \reals^{F_{T_{\rmD}}}$ and $\bbc_l \in \reals^n$ are the learnable parameters of the nonlinear layer and $\varphi_\text{relu}$ is the relu function. For notational simplicity, we use the same symbols for the nonlinear layer parameters in both the primal and dual networks, though they are trained independently. The binary mask vector $\bbm \in \{0,1\}^{n}$ is defined such that $m_i = 1$ when $i \in \ccalI$ and $0$ otherwise. The Hadamard product ensures that the multipliers corresponding to the unconstrained agents are always set to zero, while the relu function guarantees that the multipliers are always nonnegative. 

\begin{figure*}[t]
    \centering
    \includegraphics[width=\linewidth, height=0.25\linewidth]{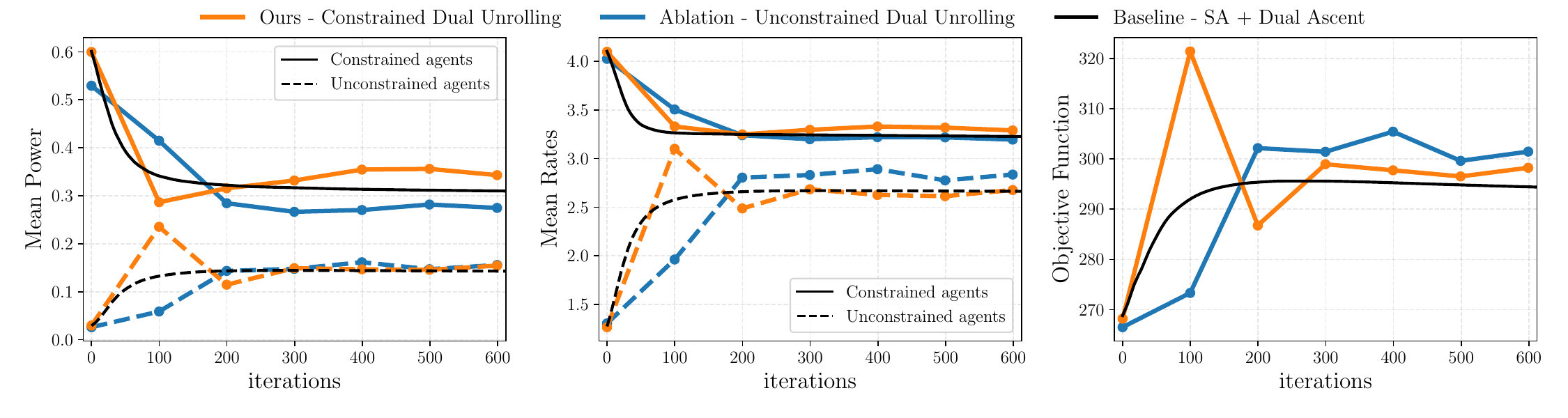}
    \vspace{-20pt}
    \caption{\small {Comparison of the mean power, rates and objective of three iterative algorithms}: Constrained Dual Unrolling (ours), its unconstrained counterpart, and state-augmented dual ascent. The 6-layer outputs of dual unrolling methods are evenly distributed across the 600 iterations for clearer visual comparison. Our constrained dual unrolled networks achieve comparable performance to dual ascent in much fewer steps.}
    \label{fig:PA_performance}
    \vspace{-8pt}
\end{figure*}

\vspace{-7pt}
\subsection{Experiment Setup}
In our numerical experiments, we consider networks of $n=100$ pairs of transmitters and receivers, deployed over a $1500 \times 1500 \ m^2$ area. The transmitters are placed randomly, and each receiver is positioned at a distance between $20$ and $60$ meters from its associated transmitter. The channel state matrix represents large-scale fading that follows a dual-slope path-loss model similar to \cite{zhang2015downlink, uslu2025generative}, combined with $7$dB log-normal shadowing. We set the maximum transmit power to $P_\text{max} = 0$ dBm, the bandwidth to $W=20$ MHz and the noise spectral density to $N_0 = -174$ dBm/Hz. In each network realization, 
$50\%$ of the transmitter-receiver pairs are randomly selected to be constrained, i.e., required to maintain a minimum rate of  $r_\text{min} = 1.5$ bps/Hz.

We deploy a primal network that consists of $K = 4$ unrolled primal layers, each comprising $T_\rmP = 3$ graph sub-layers and a dual network that consists of $L = 6$ dual layers, each also containing $T_\rmD = 3$ graph sub-layers. In both networks, each graph layer aggregates information from $K_\rmh = 2$-hop neighbors, maintains $F_l = F_k = 64$ hidden features across all intermediate layers, and deploys a leaky relu activation function.

\textbf{Training.}  For training the primal network, we use a dataset of $512$ channel state networks divided into mini-batches of a single network. At each iteration, the recent dual network is used to generate multiplier trajectories from which we select $32$ samples per network realization. We augment this set with $64$ samples drawn from a uniform distribution $\mathrm{Unif}[0, 1]$ and additional $32$ samples drawn from trajectories made by a classical dual ascent algorithm (the first benchmark below). For training the dual network, we use a dataset of $2048$ channel state networks divided into batches of $256$ graphs. The initial multipliers $\bblambda_0$ are initialized such that entries corresponding to constrained users are sampled from a uniform distribution $\mathrm{Unif}[0,10]$, while all others are set to zero. In each iteration of our joint training scheme, we perform one epoch of primal network training (Algorithm~\ref{alg:primal-training}) for every $5$ epochs of dual network training (Algorithm~\ref{alg:dual-training}). 

In training both networks, we use ADAM optimizers to update the primal and dual parameterizations with learning rates of $\epsilon_\rmP = 10^{-4}$ and $\epsilon_\rmD = 10^{-5}$, respectively. The meta multipliers are updated through gradient ascent with stepsizes $\eta_\rmP = 10^{-3}$, and $\eta_\rmD = 10^{-3}$. We share the constraint parameter $\alpha_k = 1.05$ and $\beta_l = 0.8$ across all the primal and dual layers. {The offline training time is approximately 9 hours. Each training epoch requires 181.46 s for the primal network and 8.83 s for the dual network.}

\begin{figure*}[t]
    \centering
    \includegraphics[width=\linewidth, height=0.225\linewidth]{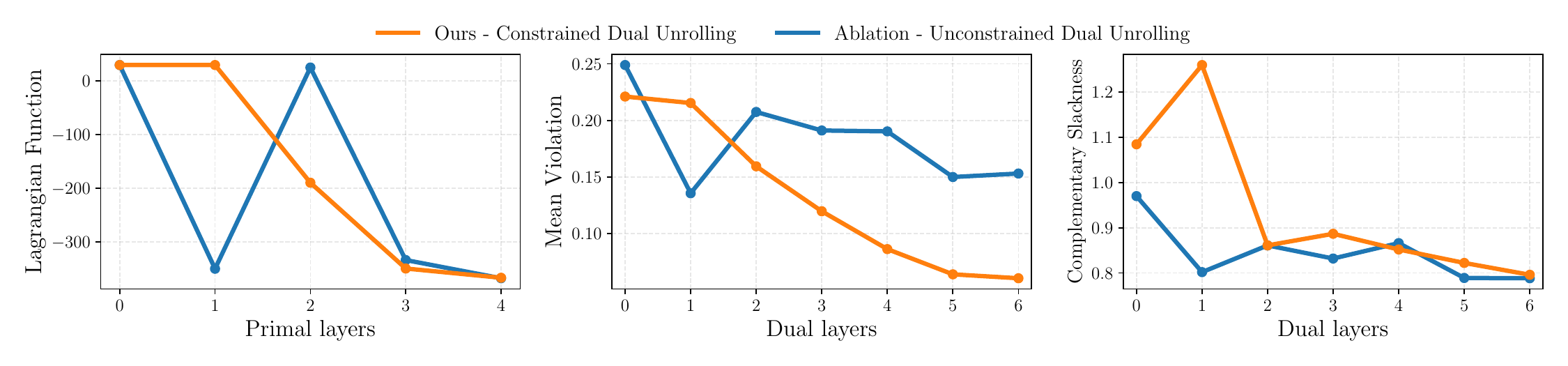}
    \vspace{-15pt}
    \caption{\small {Descent Guarantees.} (Left) The Lagrangian function across the unrolled primal layers, evaluated over test dataset of problem instances and multiplier samples. (Middle) The constraint violation across the unrolled dual layers. (Right) The complementary slackness $\bblambda_L^\top \max\{\mathbf{0}, \bbf(\bbx_l)\}$ across the unrolled dual layers. Our constrained model preserves a descent behavior over the unrolled layers, in contrast to the unconstrained model. }
    \label{fig:descent-constraints}
    \vspace{-8pt}
\end{figure*}

\textbf{Benchmarks.}  We compare the performance of our constrained dual unrolling with the following benchmarks over a testset of $128$ channel state networks:
\begin{enumerate}[label=\roman*.]
    \item State-augmentation (SA) paired with dual dynamics \cite{naderializadeh2022state}, where we use the trained primal network to predict the optimal power allocation $\bbp^\star(\bblambda_l)$ for the most recent multiplier $\bblambda_l$ and then update the multipliers using a classical gradient ascent algorithm. The iterative update is
    \begin{align}
                \bbp_l & ~=~ \bbPhi_\rmP \big(\, \bblambda_l,\,  \bbH;\, 
    \bbtheta_\rmP^\star \, \big), \nonumber \\
                \boldsymbol{\lambda}_{l+1} & ~=~ \big[ \, \boldsymbol{\lambda}_{l} + \eta \ \bbf \big(\, \bbp_l\, ; \bbH \big) \, \big]_+, \label{eq:DA_updates} \nonumber
        \end{align}
    where $\bbf$ combines the rate constraints in \eqref{eq:constrained_agents} and \eqref{eq:unconstrained_agents}.
    We set the stepsize to $\eta = 0.05$.
    \item The WMMSE-ADMM~\cite{chi2023mimo} and ITLinQ~\cite{naderializadeh2014itlinq} algorithms.
    \item A naive GNN, consisting of $12$ graph layers structured as in \eqref{eq:graph_layer2}, trained to predict the optimal power allocation directly using a supervised learning paradigm.  
    \item Unconstrained dual unrolling, where we train the same primal and dual architectures under identical hyperparameters, but without the descent constraints.
\end{enumerate}


\vspace{-7pt}
\subsection{Numerical Results and Discussions}
Figure \ref{fig:PA_performance} illustrates the iterative behavior of our constrained dual unrolling. As shown in the figure, our method follows the rate and power trajectories generated by the iterative SA method for both constrained and unconstrained users, achieving comparable performance in significantly fewer iterations. Specifically, our dual unrolling methods reach near-optimal solutions in just $6$ unrolled layers compared to $600$ iterations required by the dual ascent method it unrolls. It is important to note that both methods use the same primal network, which estimates the Lagrangian minimizer $\bbp_l \approx \bbp^\star(\bblambda_l)$ for each multiplier $\bblambda_l$. The reduction in iterations, therefore, stems from approximating the dual updates with a forward pass of the dual network. {Table~\ref{tab:runtime_target} quantifies this computational advantage. CDU reaches its target rate and violation in $0.97$ ms per problem instance, compared with $5.81$ ms for WMMSE-ADMM. Dual ascent reaches the same violation in $5.67$ ms but only jointly attains the target rate and feasibility when warm-started with our solution.}

\begin{figure*}
    \centering    
    \includegraphics[width=\linewidth]{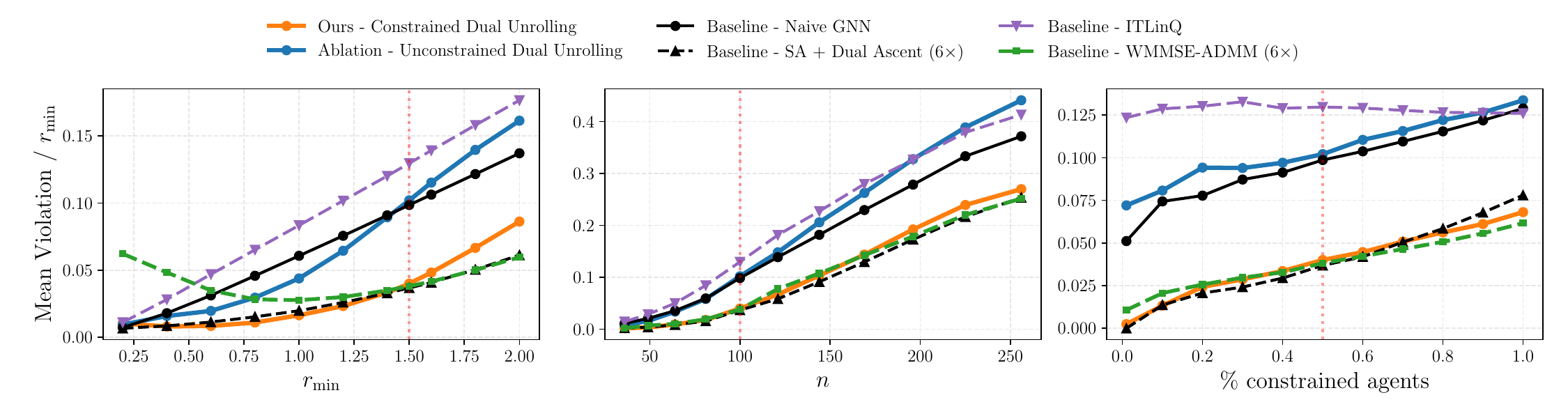}
    \vspace{-20pt}
    \caption{ \small
    {Robustness under OOD problems}, varying (left) the constraint level $r_\text{min}$, (middle) the network size $n$, and (right) the percentage of constrained agents. The red dotted line represents the in-distribution scenario: $r_\text{min}=1.5, n=100,$ and $50\%$ constrained agents. The iterative methods are given $6\times$ the inference-time budget of CDU. Across all three OOD shifts, CDU degrades more gradually than the
    unconstrained and supervised learning baselines and remains competitive with
    the iterative optimization methods.}
    \label{fig:pa_ood}
    \vspace{-10pt}
\end{figure*}

In an ablation study, we train the same primal and dual networks without descent constraints to assess their impact on performance. 
Figure \ref{fig:descent-constraints} shows that CDU preserves descent behavior across the primal and dual layers, whereas the unconstrained one fails to do so. 
Consistent with the QP example, this failure to descend results in poorer performance at the final layer. This is evidenced by a higher mean violation (0.153) compared to the constrained case (0.06).

\begin{table}[t]
\centering
\caption{\small Mean wall-clock time (ms) to reach target accuracy. The target is set to the performance of our 
$6$-layer model. Warm-start runtimes include the $0.97$ ms initialization cost.
``--'' indicates that the target is not reached.}
\label{tab:runtime_target}
\resizebox{\columnwidth}{!}{%
\begin{tabular}{lccc}
\toprule
{Method}
& {Violation $\leq 0.06$}
& {Rate $\geq 2.98$}
& {Both} \\
\midrule
{Constrained Dual Unrolling}
    & \textbf{0.97 }
    & \textbf{0.97 }
    & \textbf{0.97 } \\
SA + Dual Ascent
    & 5.67 
    & --
    & -- \\
SA + Dual Ascent (warm)
    & 1.20 
    & 1.20 
    & 1.20 \\
WMMSE-ADMM
    & 5.81 
    & 2.91 
    & 5.81  \\
WMMSE-ADMM (warm)
    & 6.80 
    & 3.88 
    & 6.80  \\
\bottomrule
\end{tabular}%
}
\vspace{-10pt}
\end{table}

{To examine the OOD performance, we test our model on datasets generated by varying a single parameter--either the constraint level $r_\text{min}$, the network size $n$, or the constrained-agent percentage. 
Figure~\ref{fig:pa_ood} shows that CDU degrades
smoothly under all three shifts. As $r_{\min}$ increases, the problem becomes progressively harder and the violation of all methods increases, but CDU remains substantially below the unconstrained ablation, the naive GNN, and ITLinQ, while remaining competitive with the iterative optimization
baselines, despite operating with one-sixth of their inference-time budget. A similar trend appears as the network size grows. The violation of CDU increases with $n$, reflecting the stronger interference in denser networks, but its degradation is slower than that of the learning-based baselines. Finally, when the fraction of constrained agents increases, CDU again exhibits a gradual increase in violation and preserves a clear advantage over its unconstrained counterpart and the naive GNN. These results indicate that the descent and ascent constraints improve not
only in-distribution feasibility, but also the stability of the learned solver as the constraint difficulty and problem scale move away from the training regime.}

\section{Conclusions}

This paper introduced an unrolling-based framework for constrained optimization that jointly trains a pair of unrolled networks to find the saddle point of the Lagrangian function. By imposing layerwise descent and ascent constraints, CDU preserves the iterative structure of primal-dual optimization while learning the update mappings themselves. We established feasibility and optimality guarantees for the coupled architecture. Across quadratic programs and wireless power-allocation problems, CDU achieves good-quality solutions with substantially lower time-to-target than classical iterative solvers, while also improving robustness to out-of-distribution changes in problem size and constraint difficulty. These results support constrained unrolling as an effective approach for learning fast optimization procedures without hard-coding the analytical updates of a specific solver.

\vspace{-8pt}
\begin{appendices}

\section{Proof of Lemma~\ref{lem:generic}}\label{app:generic}

The proof follows~\cite[Theorem 2 and Lemma 1]{hadou2024Robust} and is summarized here for completeness. Let $\mathcal{A}_k\in\mathcal{F}_k$ denote the event that the $k$th
population constraint is satisfied. By the constrained-learning
guarantee,
$
    \mathbb{P}(\mathcal{A}_k)
    =
    1-\delta
$.
Using the law of total expectation,
\begin{equation}\label{eq:total_expectation}
    \mbE[V_k]
    ~=~
    \mathbb{P}(\mathcal{A}_k)
    \mbE[V_k\mid\mathcal{A}_k]
    +
    \mathbb{P}(\mathcal{A}_k^c)
    \mbE[V_k\mid\mathcal{A}_k^c].     
\end{equation}
On $\mathcal{A}_k$, near-feasibility of the learned constraint gives
\begin{align}
    \mbE[V_k\mid\mathcal{A}_k]
    ~\leq~
    \beta\mbE[V_{k-1}]
    +
    \zeta.
\end{align}
On $\mathcal{A}_k^c$, boundedness in Assumption~\ref{as:regularity} gives
\begin{align}
    \mbE[V_k\mid\mathcal{A}_k^c]
    ~\leq~
    M.
\end{align}
Therefore,~\eqref{eq:total_expectation} is upper bounded by
\begin{align}
    \mbE[V_k]
    ~\leq~&
    (1-\delta)\beta
    \mbE[V_{k-1}]
    +
    (1-\delta)\zeta
    +
    \delta M.
    \label{eq:primal_recursion}
\end{align}

Let
$
    Z_k
    :=
    \mbE_{\bbz,\bblambda}[V_k]
$ be the expectation of $V_k$ over the training distribution over $\bbz$ and $\bblambda$.
Conditioning \eqref{eq:primal_recursion} on $\mathcal{F}_{k-1}$
yields
\begin{align}
    \mbE[Z_k & \mid\mathcal{F}_{k-1}]
    ~\leq~
    (1-\delta) Z_{k-1}
    -
    (1-\delta) (1-\beta)
    \big[
         Z_{k-1}
        -
        \eta
    \big],
    \label{eq:primal_supermartingale}
\end{align}
with $\eta$ as defined in~\eqref{eq:generic-liminf}.
This conditional distribution is taken over the noise we add at each layer.
Define
$
    Z_k^{\mathrm{best}}
    ~:=~
    \min_{j\leq k} \, \, Z_j,
$
together with
\begin{align}
    A_k
    &:=
    Z_k
    \mathbf{1} \,
    \left\{
        Z_k^{\mathrm{best}}>\eta
    \right\},
    \\
    B_k
    &:=
    (1-\beta)
    \big[
         Z_{k-1}
        -
        \eta
    \big]\,
    \mathbf{1}
    \left\{
        Z_k^{\mathrm{best}}>\eta
    \right\}.
\end{align}
By the definition of $\eta$, both $A_k$ and $B_k$ are
nonnegative whenever the indicator is active.
The first sequence keeps the values of $Z_k$ up until the best
value $Z_k^{\mathrm{best}}$
drops below $\eta$ for the first time. After this point,
$A_k$ is zero. The second sequence tracks the difference $(1-\beta)
    \big[
         Z_{k-1}
        -
        \eta
    \big]$ until $A_k$ becomes zero.
 Hence, 
 $\mbE[A_k \mid \ccalF_{k-1}, A_{k-1} \neq 0] = \mbE[Z_k \mid \ccalF_{k-1}, A_{k-1} \neq 0]$, 
 $\mbE[A_k \mid \ccalF_{k-1}, A_{k-1} = 0] = 0$, and
\begin{align}
    \mbE
    \left[
        A_k
        \mid
        \mathcal{F}_{k-1}
    \right]
    &~\leq~
    \mbE
    \left[
        Z_k
        \mid
        \mathcal{F}_{k-1}
    \right] \nonumber\\
    & ~\leq~
    (1-\delta) \Big(A_{k-1}
    -
    B_{k-1} \Big).
\end{align}
The Robbins--Siegmund supermartingale convergence theorem therefore
implies that $A_k$ converges almost surely and
$
    \sum_{k=0}^{\infty} B_k
    <
    \infty
$, a.s.
Hence
$
    B_k
    \rightarrow
    0$, \, 
    a.s.

If the indicator is extinguished at some finite layer, then
$Z_k^{\mathrm{best}}\leq\eta$ from that layer onward. Otherwise,
the indicator remains active indefinitely, and $B_k\rightarrow0$
implies
\begin{align}
    \liminf_{k\rightarrow\infty} \, 
    Z_k
    ~=~ 
    \liminf_{k\rightarrow\infty} \, 
    \mbE[V_k]
    \leq
    \eta.
\end{align}
Thus, in either case,
$
    \lim_{k\rightarrow\infty}
    Z_k^{\mathrm{best}}
    ~\leq~
    \eta$, a.e.

By Fatou's lemma, $\int \liminf_k V_k d\mu \leq \liminf_k \int V_k d\mu$, which gives
\begin{align}
    \lim_{k\rightarrow\infty}
    \mbE
    \left[
        \min_{j\leq k} V_j
    \right]
    ~\leq~
    \eta
    \qquad
    \text{a.s.}
\end{align}

For the finite-depth result, recursively applying
\eqref{eq:primal_recursion} gives
\begin{align}
    \mbE[V_K]
    ~\leq~&
        (1-\delta)^K\beta^K
    \mbE[V_0]
    \nonumber\\
    &+
    \Big[
        (1-\delta)\zeta
        +
        \delta M
    \Big]
    \sum_{j=0}^{K-1}
        (1-\delta)^j\beta^j.
\end{align} 
Bounding the geometric sum by $\big(1-(1-\delta)\beta\big)^{-1}$ and using
$1-(1-\delta)\beta \ge (1-\delta)(1-\beta)$ gives
\begin{align}
    \mbE[V_K]
    & \le (1-\delta)^K\beta^K\,\mbE[V_0]
    + \frac{(1-\delta)\zeta + \delta{M}}{(1-\delta)(1-\beta)} \nonumber \\
    & = (1-\delta)^K\beta^K\,\mbE[V_0] + \eta,
\end{align}
which completes the proof. \qed

\section{Proof of Theorem~\ref{thm:primal}}\label{app:primal}
The first claim is \eqref{eq:generic-lastlayer} of
Lemma~\ref{lem:generic} with $V_k =
\|\nabla_{\mathbf{x}}\mathcal{L}(\widetilde{\mathbf{x}}_k,\bblambda;
\mathbf{z})\|$, $\beta = {\alpha}$, ${M} = M_{\rmP}$,
$(\zeta,\delta) = (\zeta_{\rmP},\delta_{\rmP})$, where
\eqref{eq:generic-constraint} holds by Lemma~\ref{lem:constrained_learning}. The second
claim follows by repeating the argument of
\cite[Corollary~1]{hadou2024Robust}. Under the shifted distribution, each
descent constraint is satisfied up to
$\zeta_{\rmP} + 2M_{\rmP}
d(\mathcal{D}_{\bm\lambda},\mathcal{D}'_{\bm\lambda})$, rather than $\zeta_\rmP$, w.p. $1-\delta_{\rmP}$ each.
\qed

\section{Proof of Lemma~~\ref{lem:bias}}\label{app:bias}

Since $\ccalL(\cdot,\bblambda_l;\bbz)$ is
$m$-strongly convex,
\begin{equation}
0
\leq
\ccalL(\bbx_l,\bblambda_l;\bbz)
-
g(\bblambda_l;\bbz)
\leq
\frac{1}{2m}
\left\|
\nabla_{\bbx}
\ccalL(\bbx_l,\bblambda_l;\bbz)
\right\|^2.
\end{equation}
By Assumption~\ref{as:regularity},
$\|\nabla_{\bbx}\ccalL\|\leq M_{\rm P}$, and hence
\begin{equation}
\ccalL(\bbx_l,\bblambda_l;\bbz)
-
g(\bblambda_l;\bbz)
\leq
\frac{M_{\rm P}}{2m}
\left\|
\nabla_{\bbx}
\ccalL(\bbx_l,\bblambda_l;\bbz)
\right\|.
\end{equation}
Taking expectations and applying the shifted-distribution bound of
Theorem~\ref{thm:primal} gives
\begin{equation}
\mathbb{E}\!\left[
\ccalL(\bbx_l,\bblambda_l;\bbz)
-
g(\bblambda_l;\bbz)
\right]
\leq
\frac{M_{\rm P}}{2m}\rho_K.
\end{equation}
Since the positive-part operator is nonexpansive and
$\bbf(\cdot;\bbz)$ is $L_f$-Lipschitz,
\begin{align}
    \big\|[\bbf(\bbx_l;\bbz)]_+
    -
    [\bbf(\bbx^\star(\bblambda_l);\bbz)]_+\big\|
    &\leq
    \left\|
    \bbf(\bbx_l;\bbz)
    -
    \bbf(\bbx^\star(\bblambda_l);\bbz)
    \right\|
    \nonumber\\
    &\leq
    L_f
    \|\bbx_l-\bbx^\star(\bblambda_l)\|.
\end{align}
Strong convexity further gives
\begin{align}
    \|\bbx_l-\bbx^\star(\bblambda_l)\|
    \leq
    \frac{1}{m}
    \|\nabla_{\bbx}
    \ccalL(\bbx_l,\bblambda_l;\bbz)\|.
\end{align}
Taking expectations and applying Theorem~\ref{thm:primal} yields
\eqref{eq:feas_bias}.
\qed

\section{Proof of Theorem~\ref{thm:dual_convergence}}
\label{app:dual_convergence}

The feasibility guarantee follows directly from
Lemma~\ref{lem:generic}. Let
$
    V_l
    =
    \left\|
        [\bbf(\bbx_l;\bbz)]_+
    \right\|.
$
The layerwise constraints in~\eqref{eq:empirical_dual_training} impose the
corresponding contraction with rate $\beta$, up to the
constrained-learning error $\zeta_D$ and with probability
$1-\delta_D$. Since $V_l\leq M_D$ by Assumption~\ref{as:regularity},
Lemma~\ref{lem:generic} gives~\eqref{eq:dual_feasibility_bound}.

We next establish dual optimality 
by combining the near-optimality guarantee of Lemma~\ref{lem:constrained_learning} with the primal approximation bound of Lemma~\ref{lem:bias}.
We relate $P_{D,\mathrm{fun}}^\star$ to the true dual optimum. Consider the
oracle multiplier mapping $\boldsymbol\lambda^\star(\mathbf z)$, and let
\begin{equation}
\bar{\mathbf x}^\star
=
\Phi_P(\boldsymbol\lambda^\star,\mathbf z;\boldsymbol\theta_P^\star)
\end{equation}
denote the corresponding primal-network output. Since the exact primal solution
$\mathbf x^\star(\boldsymbol\lambda^\star)$ is feasible, i.e.,
$
[\mathbf f(\mathbf x^\star(\boldsymbol\lambda^\star);\mathbf z)]_+~=~0.
$
Lemma~\ref{lem:bias} therefore gives
\begin{equation}
\mathbb E\!\Big[
\|[\mathbf f(\bar{\mathbf x}^\star;\mathbf z)]_+\|
\Big]
\le b_f.
\label{eq:oracle_primal_feas}
\end{equation}

Consider now the constant functional trajectory
\begin{equation}
\boldsymbol\lambda_0
=
\cdots
=
\boldsymbol\lambda_L
=
\boldsymbol\lambda^\star.
\end{equation}
For this trajectory, each layerwise training constraint has residual
\begin{align*}
\mathbb E \Big[
[\mathbf f(\bar{\mathbf x}^\star;\mathbf z)]_+
-
\beta
[\mathbf f(\bar{\mathbf x}^\star;\mathbf z)]_+
\Big]
=
(1-\beta)
\mathbb E\Big[
[\mathbf f(\bar{\mathbf x}^\star;\mathbf z)]_+
\Big].
\end{align*}
By \eqref{eq:oracle_primal_feas}, this trajectory is feasible for a relaxation
of the functional dual-training problem by at most $(1-\beta)b_f$ in each
constraint.
Let $P_{D,\mathrm{fun}}^\star(r)$ denote the optimum of this relaxed
functional problem, with
$P_{D,\mathrm{fun}}^\star(0)=P_{D,\mathrm{fun}}^\star$.
Since the oracle trajectory is feasible for the relaxed problem,
\begin{align}
P_{D,\mathrm{fun}}^\star(r)
&\ge
\mathbb E\!\left[
\mathcal L(\bar{\mathbf x}^\star,
\boldsymbol\lambda^\star;\mathbf z)
\right] \ge
\mathbb E[g^\star(\mathbf z)],
\end{align}
where the second inequality follows from
$g^\star(\mathbf z)
=\min_{\mathbf x}
\mathcal L(\mathbf x,\boldsymbol\lambda^\star;\mathbf z)$.

Let $\boldsymbol\xi^\star$ denote an optimal multiplier associated with the
constraints of the unrelaxed functional problem. By Assumption~\ref{as:clt} and the
functional strong-duality result of constrained learning
\cite[Proposition~III.2]{chamon2022constrained}, such a multiplier exists and
the functional primal and dual optimal values coincide. Therefore, the
standard sensitivity bound for a relaxation by $r$ gives
\begin{equation}
P_{D,\mathrm{fun}}^\star(r)
-
P_{D,\mathrm{fun}}^\star
\le
r\|\boldsymbol\xi^\star\|_1 .
\end{equation}
Combining the previous two inequalities yields
\begin{equation}
\mathbb E[g^\star(\mathbf z)]
\le
P_{D,\mathrm{fun}}^\star
+
r\|\boldsymbol\xi^\star\|_1,
\end{equation}
and therefore, substituting $r=(1-\beta)b_f$,
\begin{equation}
\mathbb E[g^\star(\mathbf z)]
-
P_{D,\mathrm{fun}}^\star
\le
(1-\beta)\|\boldsymbol\xi^\star\|_1 b_f.
\label{eq:functional_to_true_dual}
\end{equation}

Using Lemma~\ref{lem:constrained_learning} and \eqref{eq:functional_to_true_dual},
\begin{align}
&
\mathbb E[g^\star(\mathbf z)]
-
\mathbb E[
\mathcal L(\mathbf x_L,\boldsymbol\lambda_L;\mathbf z)]
~\leq~
\epsilon_D
+
(1-\beta)\|\bbxi^\star\|_1 b_f.
\label{eq:lagrangian_gap_bound}
\end{align}

Finally, add and subtract the final-layer Lagrangian:
\begin{align}
\mathbb E\!\left[
g^\star(\mathbf z)
-
g(\boldsymbol\lambda_L;\mathbf z)
\right]
=~ & 
\mathbb E\!\left[
g^\star(\mathbf z)
-
\mathcal L(\mathbf x_L,\boldsymbol\lambda_L;\mathbf z)
\right]
\nonumber\\
&+
\mathbb E\!\left[
\mathcal L(\mathbf x_L,\boldsymbol\lambda_L;\mathbf z)
-
g(\boldsymbol\lambda_L;\mathbf z)
\right].
\end{align}
The first term is bounded by \eqref{eq:lagrangian_gap_bound}, while Lemma~\ref{lem:bias} bounds the second term,
which completes the proof.
\qed

\end{appendices}

\bibliographystyle{ieeetr}
\bibliography{Bib}

\end{document}